%% file: bare_jrnl.tex
\documentclass[journal]{IEEEtran}

\usepackage{cite}
\usepackage{balance}
\usepackage{makecell}
\usepackage[affil-it]{authblk} 
\usepackage{etoolbox}
\usepackage{lmodern}
\usepackage[hyphens]{url}
\usepackage{color}
\usepackage{verbatim}
\usepackage{siunitx}
\usepackage{multicol}
\usepackage{multirow}
\usepackage{subfigure}
\usepackage{tabularx}
\usepackage{booktabs}
\usepackage{graphicx}
\usepackage{tablefootnote}
\usepackage{comment}
\usepackage{graphics}

\usepackage{algorithm}
\usepackage{algpseudocode}

\usepackage{flushend}

\usepackage[normalem]{ulem}
\usepackage{color,soul}
\usepackage{pgfplots} 
\pgfplotsset{compat=1.14}
\usepackage{graphicx}
\usepackage{threeparttable}

\newcommand{\edit}[1]{#1}

\begin{document}

\title{RANC: Reconfigurable Architecture for Neuromorphic Computing}

\author{Joshua Mack,~\IEEEmembership{}
        Ruben Purdy,~\IEEEmembership{}
        Kris Rockowitz,~\IEEEmembership{}
        Michael Inouye,~\IEEEmembership{}
        Edward Richter,~\IEEEmembership{}
        Spencer Valancius,~\IEEEmembership{}

        Nirmal Kumbhare,~\IEEEmembership{}
        Md Sahil Hassan,~\IEEEmembership{}
        Kaitlin Fair,~\IEEEmembership{}
        John Mixter,~\IEEEmembership{}

        and~Ali Akoglu~\IEEEmembership{}%

}

\maketitle

\input{sections_revised/0_abstract}


\IEEEpeerreviewmaketitle

\section{Introduction}\label{sec:introduction}
\input{./sections_revised/1_introduction.tex}

\section{RANC Ecosystem}\label{sec:ecosystem}
\input{./sections_revised/2_ecosystem.tex}

\section{Architectural Verification} \label{sec:verification}
\input{./sections_revised/3_verification.tex}

\section{Case Studies}\label{sec:case_studies}
\input{./sections_revised/4_case_studies.tex}

\section{Related Work}\label{sec:related_work}
\input{./sections_revised/5_related_work.tex}

\section{Conclusions and Future Work}\label{sec:conclusion}
\input{./sections_revised/6_conclusion.tex}

\section*{Acknowledgment}
This work is partly supported by National Science Foundation research project CNS-1624668 %
and Raytheon Missile Systems (RMS) under the contract 2017-UNI-0008. 
The content is solely the responsibility of the authors and does not necessarily represent the official views of RMS.

\bibliographystyle{IEEEtran}
\bibliography{IEEEabrv,references}

\newpage

\begin{IEEEbiography}[{
\vspace*{-13mm}\includegraphics[width=0.9in,height=0.9in,clip,keepaspectratio]{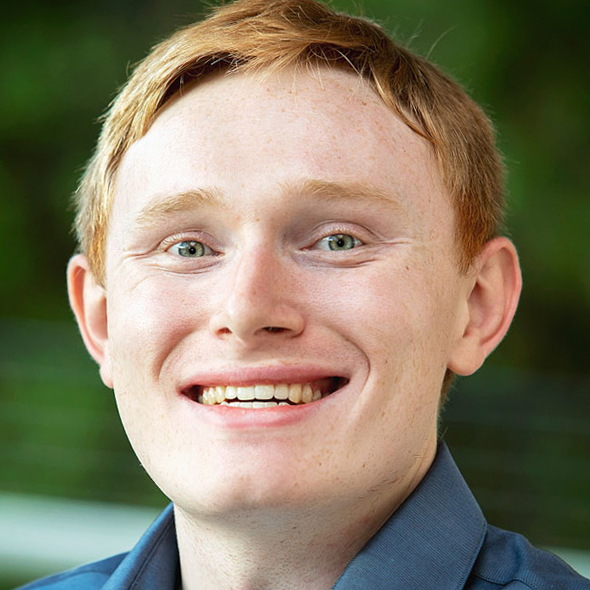}
}]{Joshua Mack}
is a Ph.D. student in the Electrical and Computer Engineering program at the University of Arizona. His research interests include reconfigurable systems; emerging architectures; and intelligent workload partitioning across heterogeneous systems.
\end{IEEEbiography}

\vskip -4.0\baselineskip plus -1fil

\begin{IEEEbiography}[{
\vspace*{-13mm}\includegraphics[width=0.9in,height=0.9in,clip,keepaspectratio]{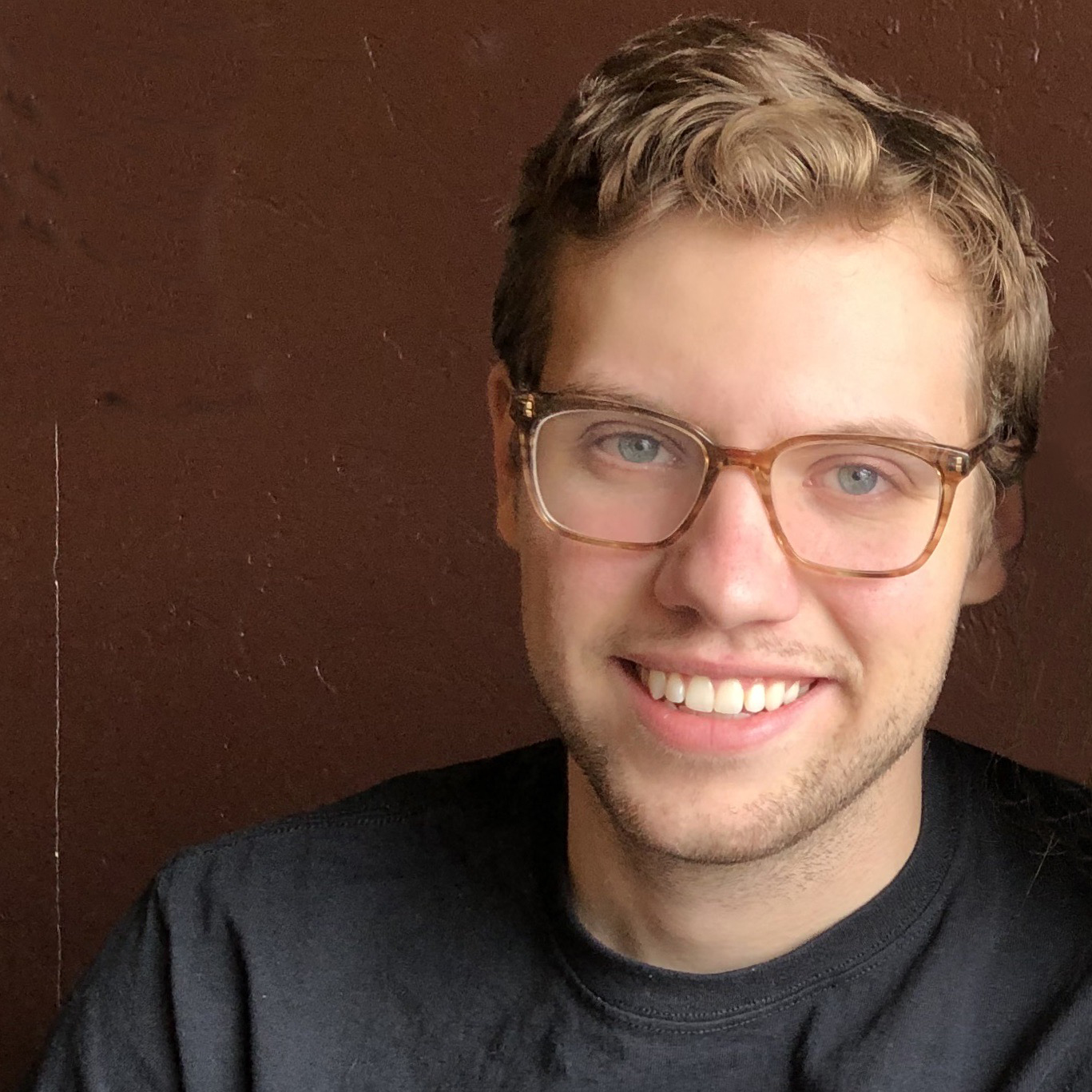}
}]{Ruben Purdy}
is a Ph.D. student in the Electrical and Computer Engineering department at Carnegie Mellon University. He received his B.S. in ECE at the University of Arizona. His research interests include energy-efficient machine learning and hardware security.
\end{IEEEbiography}

\vskip -4.0\baselineskip plus -1fil

\begin{IEEEbiography}[{
\vspace*{-13mm}\includegraphics[width=0.9in,height=0.9in,clip,keepaspectratio]{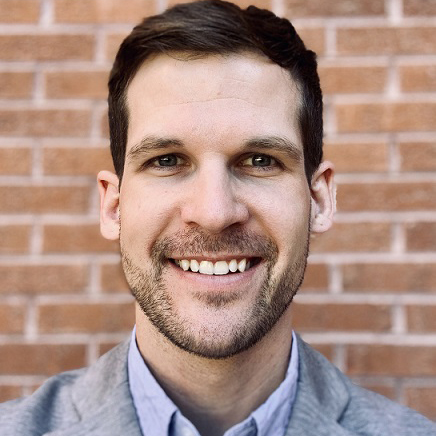}
}]{Kris Rockowitz}
is a graduate student in the Department of Electrical and Computer Engineering at the University of Arizona. Research interests include Neuromorphic Engineering, Neural Network Architectures, and High Performance Computing.
\end{IEEEbiography}

\vskip -4.0\baselineskip plus -1fil

\begin{IEEEbiography}[{
\vspace*{-13mm}\includegraphics[width=0.9in,height=0.9in,clip,keepaspectratio]{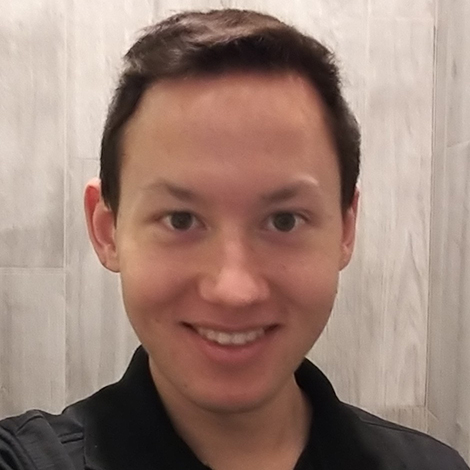}
}]{Michael Inouye}
is a graduate student in the Department of Electrical and Computer Engineering at the University of Arizona, where he received his B.S. in ECE as well. His research interests include neuromorphic computing and digital design.
\end{IEEEbiography}

\vskip -4.0\baselineskip plus -1fil

\begin{IEEEbiography}[{
\vspace*{-13mm}\includegraphics[width=0.9in,height=0.9in,clip,keepaspectratio]{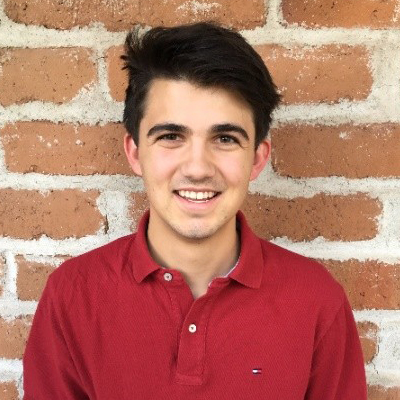}
}]{Edward Richter}
is a Ph.D. student in the Electrical and Computer Engineering department at the University of Illinois at Urbana-Champaign. He is interested in reconfigurable computing platforms for acceleration and enabling architectural and system-wide research
\end{IEEEbiography}

\vskip -4.0\baselineskip plus -1fil

\begin{IEEEbiography}[{
\vspace*{-13.5mm}\includegraphics[width=0.9in,height=0.9in,clip,keepaspectratio]{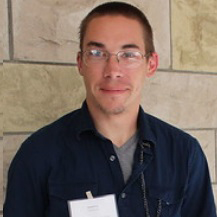}
}]{Spencer Valancius}
is a Masters graduate of the College of Electrical and Computer Engineering from the University of Arizona. He currently works at a National Laboratory in the state of New Mexico.
\end{IEEEbiography}

\vskip -4.0\baselineskip plus -1fil

\begin{IEEEbiography}[{
\vspace*{-13mm}\includegraphics[width=0.9in,height=0.9in,clip,keepaspectratio]{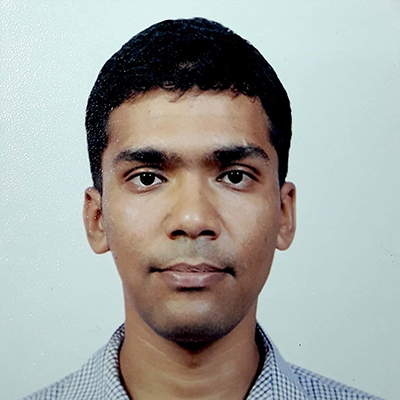}
}]{Nirmal Kumbhare}
received his Ph.D. in computer engineering from the University of Arizona. His research interests involve reconfigurable and heterogeneous systems, high performance computing, and power-aware resource management. 
He has worked with Intel India prior to joining Ph.D.
\end{IEEEbiography}

\vskip -4.0\baselineskip plus -1fil

\begin{IEEEbiography}[{
\vspace*{-13mm}\includegraphics[width=0.9in,height=0.9in,clip,keepaspectratio]{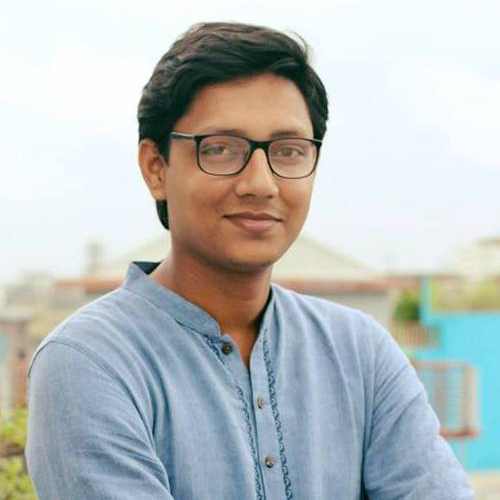}
}]{Sahil Hassan}
is a Ph.D. student in the Electrical and Computer Engineering department at the University of Arizona. His research interests include reconfigurable computing and adaptive hardware architecture design.
\end{IEEEbiography}

\vskip -4.0\baselineskip plus -1fil

\begin{IEEEbiography}[{
\vspace*{-4mm}
\includegraphics[width=0.95in,height=1.175in,clip,keepaspectratio]{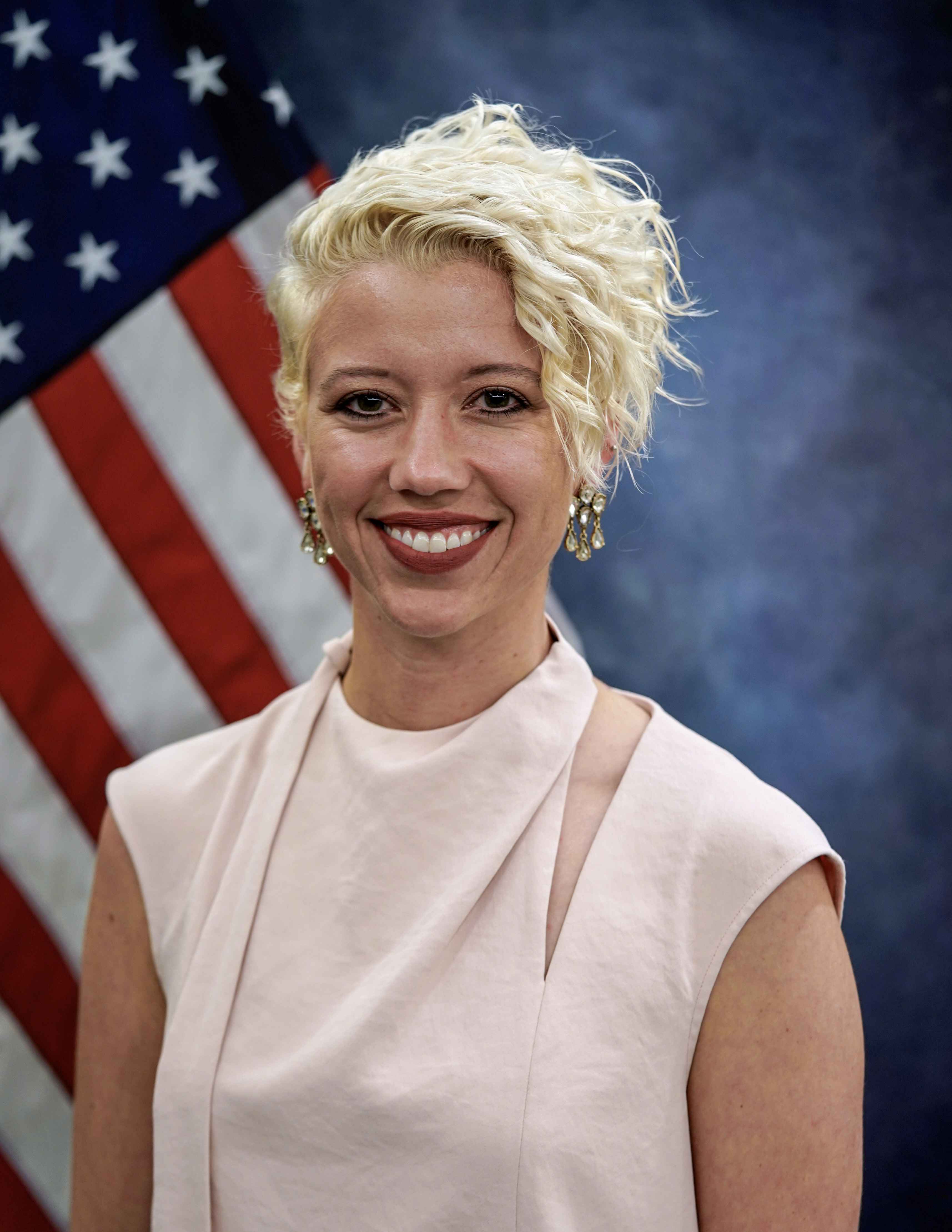}
}]{Kaitlin Fair}
Kaitlin Fair completed her PhD in Electrical Engineering at the Georgia Institute of Technology in 2017. 
She is the Integrated Sensor and Navigation Services Team Lead for the Integrated Seekers and Processing Branch at AFRL. 
Her research interests include efficient signal processing, algorithm development on brain-inspired (neuromorphic) engineering architectures, and explainable artificial intelligence. 
\end{IEEEbiography}

\vskip -2.5\baselineskip plus -1fil

\begin{IEEEbiography}[{
\vspace*{-13mm}\includegraphics[width=0.9in,height=0.9in,clip,keepaspectratio]{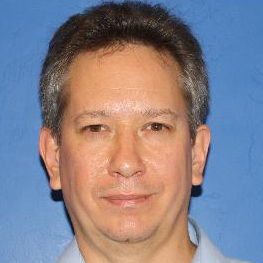}
}]{John Mixter} is employed by Raytheon Technologies regarding emerging technologies on neuromorphic engineering and AI/ML.  He is a University of Arizona PhD Candidate focusing on dynamic neural network growth within embedded hardware environments.   
\end{IEEEbiography}

\vskip -4.5\baselineskip plus -1fil
\begin{IEEEbiography}[{
\vspace*{-0mm}
\includegraphics[
width=1in,height=1.25in,clip,keepaspectratio
]{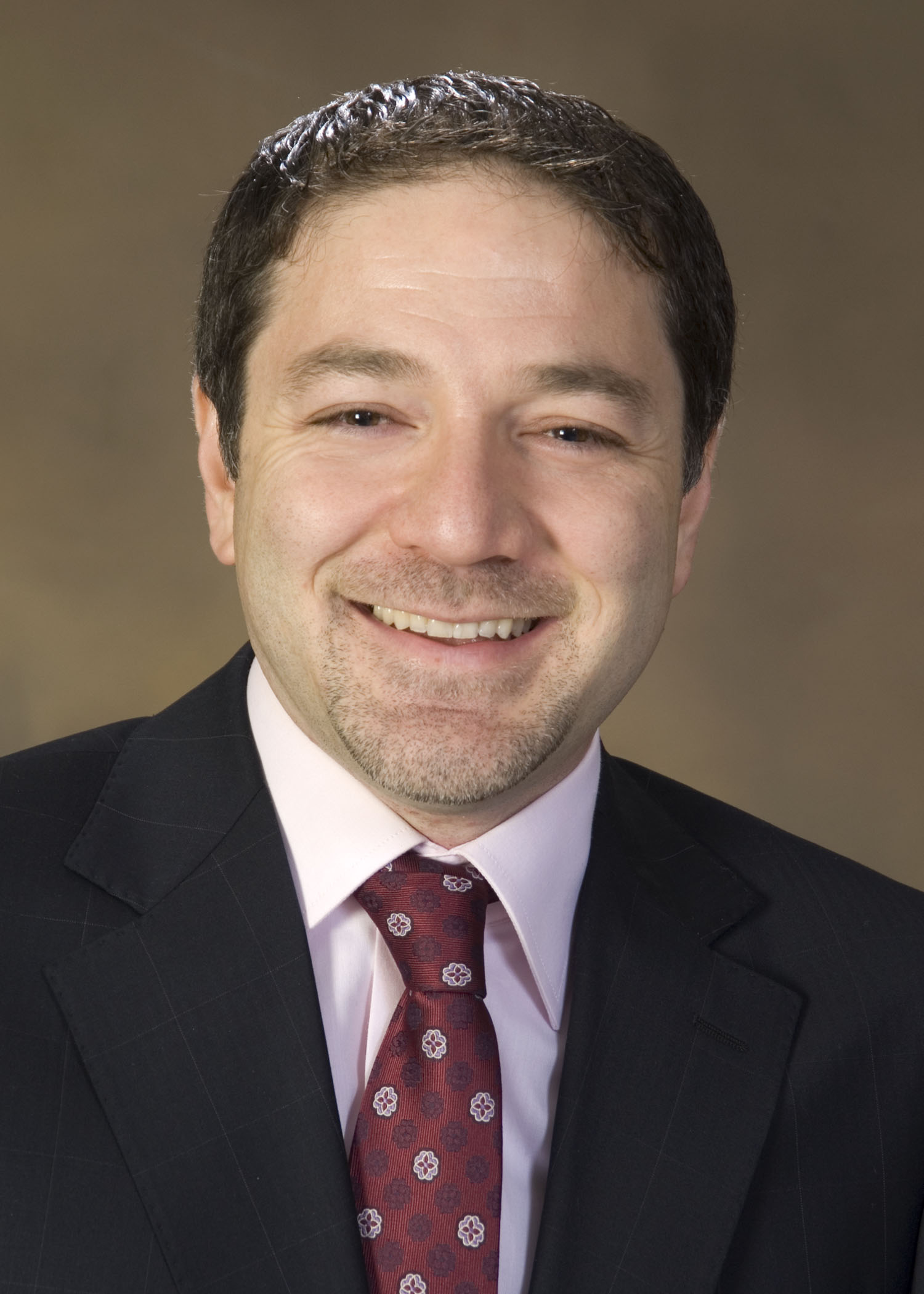}}] {Ali Akoglu} received his Ph.D. degree in Computer Science from the Arizona State University in 2005. He is an Associate Professor in the Department of Electrical and Computer Engineering and the BIO5 Institute at the University of Arizona. He is the site-director of the National Science Foundation Industry-University Cooperative Research Center on Cloud and Autonomic Computing. His research focus is on high performance computing and non-traditional computing architectures.  
\end{IEEEbiography}

\newpage

\appendices
\input{./sections_revised/9_appendices}

\end{document}

%% file: sections_revised/0_abstract.tex
\begin{abstract}

Neuromorphic architectures have been introduced as platforms for energy efficient spiking neural network execution. The massive parallelism offered by these architectures \edit{has} also triggered interest from \edit{non-machine learning} application domains. In order to lift the barriers to entry for hardware designers and application developers 
we present RANC: a \textbf{R}econfigurable \textbf{A}rchitecture for \textbf{N}euromorphic \textbf{C}omputing, an open-source highly flexible ecosystem that enables rapid experimentation with neuromorphic architectures in both software via C++ simulation and hardware via FPGA emulation. 
We present the utility of the RANC ecosystem by showing its ability to recreate behavior of the IBM’s TrueNorth and validate with direct comparison to IBM's Compass simulation environment and published literature.
RANC allows optimizing architectures based on application insights as well as prototyping future neuromorphic architectures that can support new classes of applications entirely.  We demonstrate the highly parameterized and configurable nature of RANC by studying the impact of architectural changes on improving application mapping efficiency with quantitative analysis based on Alveo U250 FPGA. We present post routing resource usage and throughput analysis across implementations of Synthetic Aperture Radar classification and Vector Matrix Multiplication applications, and demonstrate a neuromorphic architecture that scales to emulating 259K distinct neurons and 73.3M distinct synapses. 

\end{abstract}

%% file: sections_revised/1_introduction.tex
\IEEEPARstart{I}n recent years, neuromorphic computing architectures have prompted interest from numerous disciplines due to their implications in understanding biological brain behavior as well as their efficiency for deploying machine learning algorithms~\cite{arXiv_2017_survey}.
Neuromorphic computing architectures are non-von Neumann architectures that exploit the strengths of biologically inspired neural networks and couple together massively parallel computations with low power execution.
\edit{These aspects have been demonstrated successfully for applications such as data classification~\cite{arXiv_2017_OtherPlatforms, PNAS_2016_Applications, CF_2016_Applications, FrontComputNeurosci_2015_Applications, IJCNN_2016_Applications, ISCAS_2016_Applications}, LIDAR and control applications~\cite{Plank2019TENNLab}, optimization problems~\cite{Aimone2019Dynamic, tang_sparse_2017}, signal processing~\cite{Rajendran2019lowpower}, and even floating-point arithmetic~\cite{George2019IEEE}.}

In particular, IBM's TrueNorth~\cite{Akopyan2015TrueNorth} and Intel's Loihi~\cite{Davies2018Loihi} are canonical examples of architectures that utilize so-called ``leaky-integrate-and-fire" (LIF) neurons as their baseline neuron model.
On top of this, each platform implements unique features such as stochastic neuron behavior in TrueNorth or multicast routing in Loihi, but the fundamental behavior is primarily influenced by the underlying LIF neuron models.
In designing architectures for use in neuromorphic computing, there are an incredibly large number of configuration parameters -- such as number and precision of weights per neuron, neuron and axon counts per core, network topology, and neuron behavior -- that are not fundamentally inherent to the architecture but rather a product of physical constraints of architectural design.
To enable productive research in these kinds of architectures, there is a need for an open-source, configurable emulation environment where hardware engineers and application designers can investigate performance bottlenecks and explore design optimizations that stress evaluation of their full hardware-software stack before committing to silicon.

Fundamentally, such research is not possible with existing commercial solutions as all existing chips are presented to the end-user as pre-fabricated ASICs, and
the programmer is restricted to the architecture constraints when mapping emerging applications. 
These restrictions impact the academic research community's ability to explore novel ideas without absorbing the costs associated with manufacturing custom silicon of their own.
As such, the lack of open-source neuromorphic architecture emulation tools has a dampening effect on the research community's ability to converge towards the next generation of neuromorphic architectures.
As a point of comparison, FPGA architectures have continuously evolved, since their inception, from homogeneous island-style, 2-input lookup table-based designs to heterogeneous, 6-input lookup table-based designs with DSP and %
memory blocks to meet the needs of the ever growing application needs.  
Open source tools~\cite{rose_vtr_2012} have played a key role in this progress. 
We envision that neuromorphic computing is at a similar crossroads, with a large investment in architectural research emerging as we continue to reach the limits of transistor scaling.
To facilitate the development of future neuromorphic platforms, there is a need for a cohesive suite that enables easy and rapid prototyping of novel architectures.
There is a large body of work on neuromorphic architecture research through software based simulation environments that
support a rich set of biological features~\cite{fardet2020NEST,stimberg2019brian2,carnevale2006NEURON,Chou2018CARLSim}.
These software simulation tools are coupled with a strong drive for compiler designs~\cite{song2020compiling} that can effectively restructure applications and enable maximizing resource usage under prefabricated hardware constraints.
We approach this area from the complementary perspective of designing reconfigurable neuromorphic architectures that can be adapted to a variety of application requirements and enable rich research to be conducted on fundamental questions about applications and architectures for neuromorphic computing.
To this end, in this study we propose the following contributions:

\begin{itemize}
    \item We introduce RANC: a \textbf{R}econfigurable \textbf{A}rchitecture for \textbf{N}euromorphic \textbf{C}omputing. RANC is an open source software and hardware ecosystem\footnote{Source available at https://ua-rcl.github.io/RANC} that seeks to make neuromorphic architectures widely accessible to researchers and application developers through a cohesive programming and testing environment.
    \item We establish a baseline reconfigurable architecture, and we demonstrate RANC's configurability by recreating TrueNorth behaviorally in both C++ simulation and FPGA emulation environments through application mapping studies covering 
    classification of MNIST and EEG images, and vector matrix multiplication (VMM) execution.
    \item We reveal application-specific bottlenecks arising from architectural constraints that require application developers to take inefficient implementation approaches through architecture-specific workarounds and demonstrate \edit{this claim} with case studies on VMM and convolutional neural network execution.
    \item Driven by these application bottlenecks, we propose architectural modifications that span neuron behavior and hardware configuration changes and exemplify the capability of RANC to enable rich architectural trade studies in the area of neuromorphic computing. Together, these changes demonstrate the massive efficiencies to be gained in exploring alternative or heterogeneous neuromorphic architectures.
\end{itemize}

A preliminary version of this work appeared in the \textit{``Reconfigurable Architectures Workshop (RAW 2020)"}~\cite{RAW20} where we presented a baseline FPGA emulation platform implemented on a Xilinx Zynq Ultrascale+ ZCU102 with a focus on FPGA specific design decisions and evaluated its ability to recreate TrueNorth via case studies involving MNIST and VMM. 
In this paper, we expand the preliminary work with the following contributions:

\begin{itemize}
    \item We expand RANC to a full-stack neuromorphic research platform by integrating a software simulation and library stack on top of the previous emulator. 
    \item We develop software and hardware support in the RANC ecosystem for seamless deployment 
    onto Xilinx Alveo datacenter-scale FPGAs.
    \item We present a detailed architectural discussion on RANC's ability to replicate TrueNorth.
    \item %
    We extensively verify previous results regarding recreation of TrueNorth in RANC by extending MNIST verification to 9 and 30 core network based implementations and expanding this verification to include EEG image classification.
    \item We improve evaluations on core RANC components through cycle-by-cycle analysis of VMM execution, expose 
    and enable optimizations that are only possible with architectural customization.
    \item \edit{We demonstrate the utility of RANC in exploring architectures with heterogeneous crossbar configurations for convolution using Synthetic Aperture Radar (SAR) and CIFAR-10 image classifications.}
    \item We thoroughly explore scalability on the Alveo U250 and demonstrate RANC's ability to emulate in hardware as many as 259K neurons and 73.3M synapses.
    \item We present an exhaustive look at other neuromorphic computing environments and discuss RANC's role. 
\end{itemize}

In the next section, we will present an overview of each component in the RANC architecture, and we will discuss usage of the simulation and emulation environments.

%% file: sections_revised/2_ecosystem.tex
\subsection{Architectural Components}\label{subsec:architecture_components}

As neuromorphic architectures are designed to mimic the behavior of biological neurons in the brain, the primary data unit is the ``spike", which is typically modeled as a Dirac delta function occurring as a part of a time series.
In digital neuromorphic architectures, spikes are usually encoded as either a digital 1 or 0.
These spikes are sent to neuron units that then react accordingly by either producing more spikes or remaining dormant based on configuration parameters.
Individual neuron units communicate by sending each other spikes, and in this way, computation may be performed over time by sending carefully crafted sequences of spikes into a richly connected network of neuron units and observing the resulting behavior.

\input{sections_revised/tables/glossary.tex}

\begin{figure}[tb]
    \centering
    \includegraphics[width=0.95\linewidth]{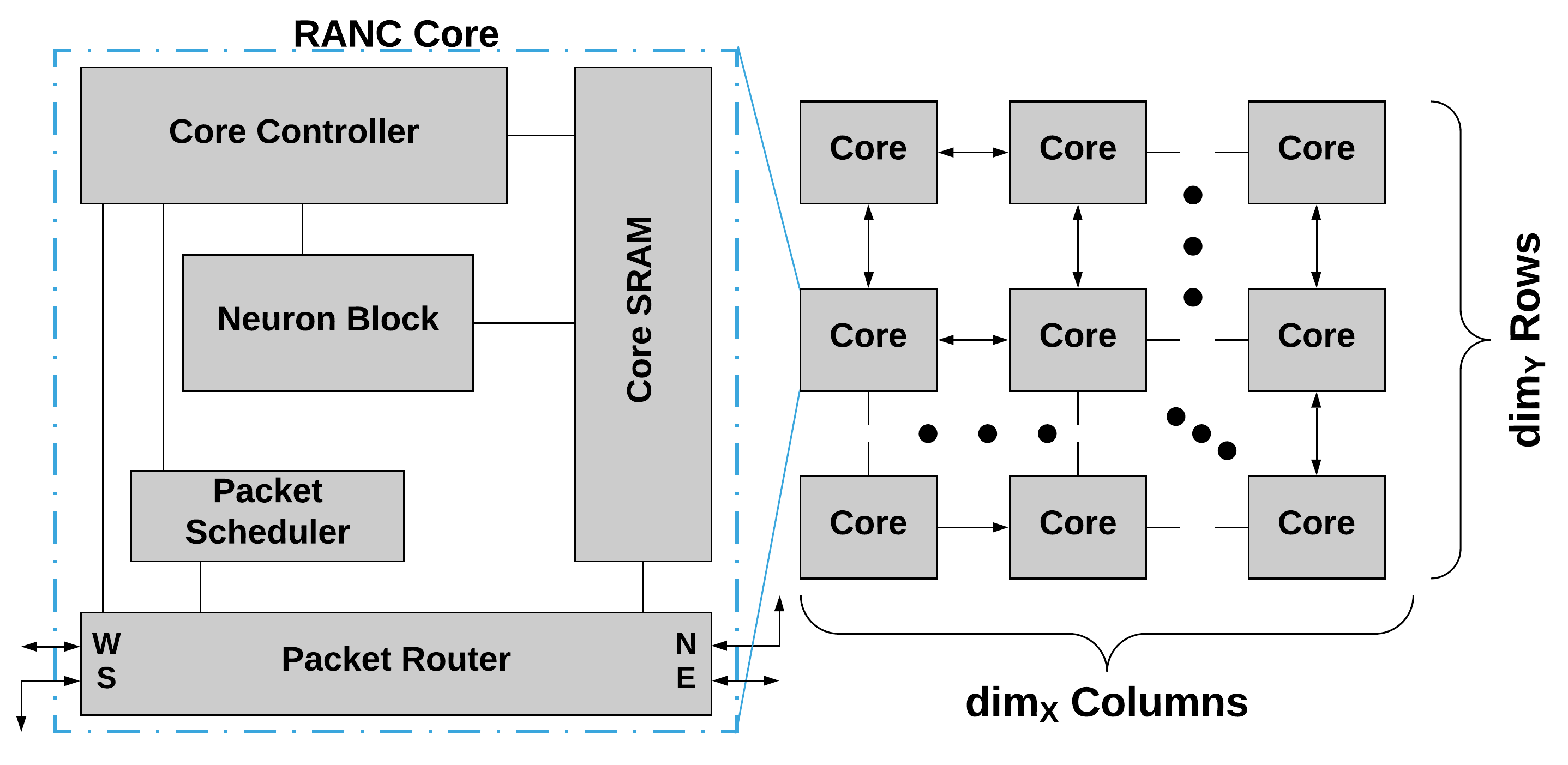}
    \caption{\edit{High level architectural overview of RANC components.}}
    \vspace{-12pt}
    \label{fig:arch_component_overview}
\end{figure}

The RANC environment, with its software simulation and hardware emulation flow,  supports the key operations of neuromorphic architectures and is highly parameterized with configurable components allowing application engineers and hardware architects to experiment with application mapping and hardware tuning concurrently.
The overall platform \edit{shown in Figure~\ref{fig:arch_component_overview}} is a 2D mesh-based network-on-chip (NoC) composed of cores with five key components: \emph{neuron block}, \emph{\edit{core controller}}, \emph{\edit{core SRAM}}, \emph{packet router}, and \emph{packet scheduler}.
\edit{This set of five components is derived by generalizing and parameterizing components of IBM's TrueNorth~\cite{Akopyan2015TrueNorth} architecture in an effort to ensure RANC provides a well-tested set of baseline neuromorphic functionalities.}
The fundamental neuron computational behaviors are implemented via the \emph{neuron block}, and each \emph{neuron block} is coupled to a \emph{\edit{core controller}} and \emph{\edit{core SRAM}} that coordinates data transfers and stores configuration parameters respectively.
Output spikes from each neuron are routed between \emph{neuron blocks} via the \emph{packet router}, and incoming spikes are scheduled for computation via the \emph{packet scheduler}.
\begin{figure}[tb]
    \vspace{-1mm}
    \centering
    \includegraphics[width=0.775\linewidth]{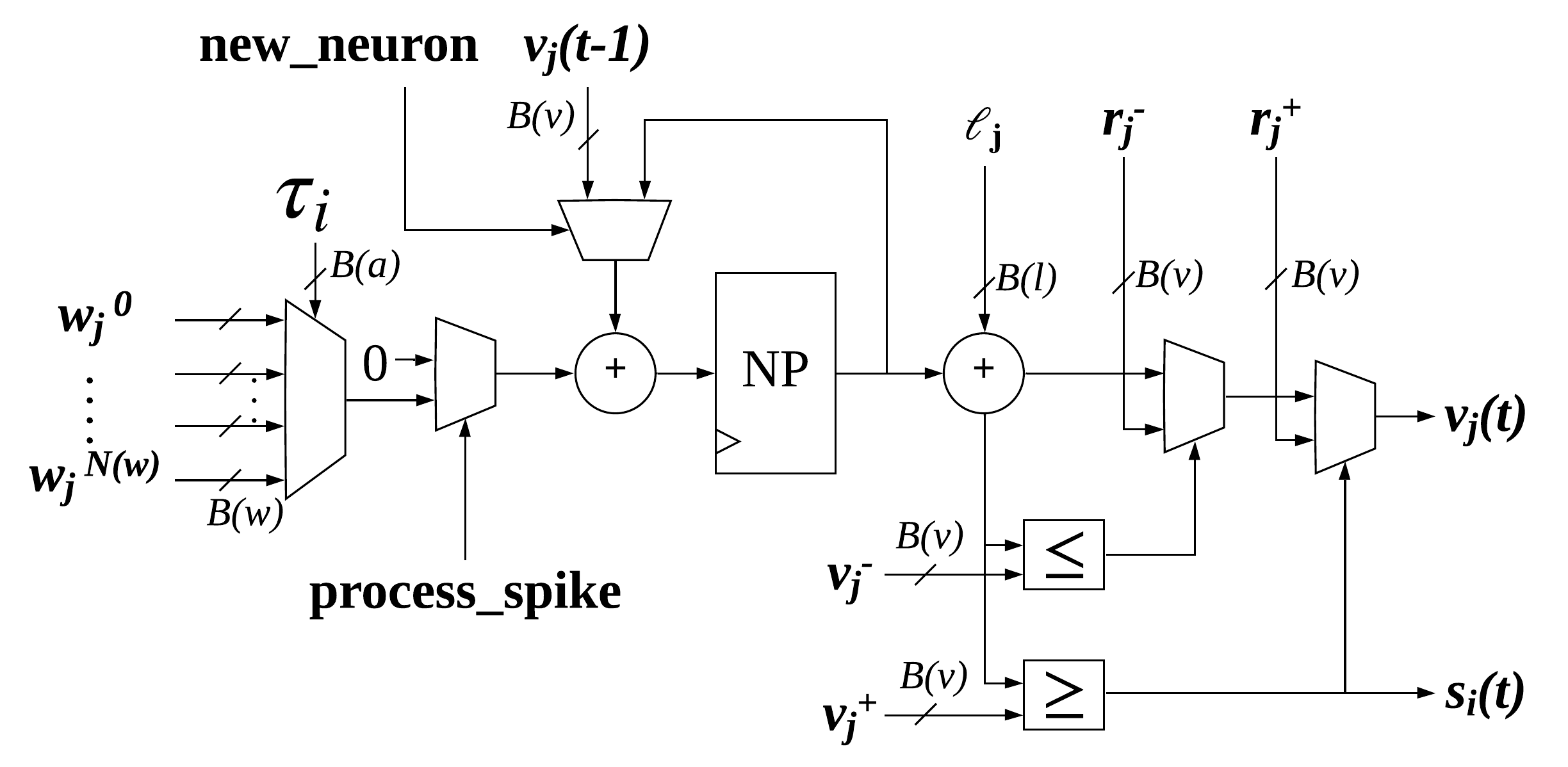}
    \vspace{-2mm}
    \caption{Design of the neuron block used in RANC.}
    \label{fig:neuron_block}
    \vspace{-12pt}
\end{figure}
Design-level synchronization is accomplished via a global synchronization signal or ``tick" that ensures all cores stay in lockstep throughout their computations.
In this way, these five core components are able to work together to form a foundational basis for neuromorphic computing architectural research while leaving flexibility to alter the fine-grained behavior of each unit independently from the rest of the design.
In the following subsections, we will discuss each component in detail.
All design parameters associated with RANC are defined in Table~\ref{tab:glossary}.
We begin by discussing the functionality of a single core whose behavior is captured by the interactions of the \emph{neuron block}, \emph{\edit{core controller}}, and \emph{\edit{core SRAM}}. We then discuss inter-core communication through the \emph{packet router} and \emph{packet scheduler}.

\subsubsection{Neuron Block}

As the primary computational component, the \emph{neuron block} emulates a crossbar with $N(a)$ input ``axons" connected to $N(n)$ output ``neurons", where $N(a)$ and $N(n)$ are parameters specified by the user.
With this, a single \emph{neuron block} can emulate a total of $N(a)\times N(n)$ synaptic connections.
Additionally, each input axon is associated with a hardcoded index that specifies the location of its associated weight value within the \emph{\edit{core SRAM}}.
To emulate each of these neurons, each \emph{neuron block} contains a basic datapath for mimicking the voltage characteristics of an LIF neuron model.
At a high level, this datapath shown in Figure~\ref{fig:neuron_block} works by having each neuron maintain a \edit{signed} running sum known as its \textit{neuron potential (NP)}. 
In each cycle, this neuron potential is either incremented, decremented, or maintained based on a number of potential events.
For each input spike that a neuron receives on one of its axons, the \edit{signed} weight associated with that axon is accumulated with the current neuron potential. 
If, at the end of this process, a neuron is above its user defined maximum threshold, then that neuron outputs a spike to its connections and the potential either resets to a static value or subtracts a user-defined value.
If, alternatively, a neuron is below its user defined minimum threshold, that neuron resets similarly either back to a static value or it adds a user-defined value to the potential without emitting a spike.
Finally, if the neuron value does not cross either the minimum or maximum threshold, it simply ``leaks" or decays by a user-specified value.
At the end of its computation, a neuron's final potential value is written into the \emph{\edit{core SRAM}}.

\subsubsection{\edit{Core Controller}}
\begin{figure}[t]
    \centering
    \includegraphics[width=0.75\linewidth]{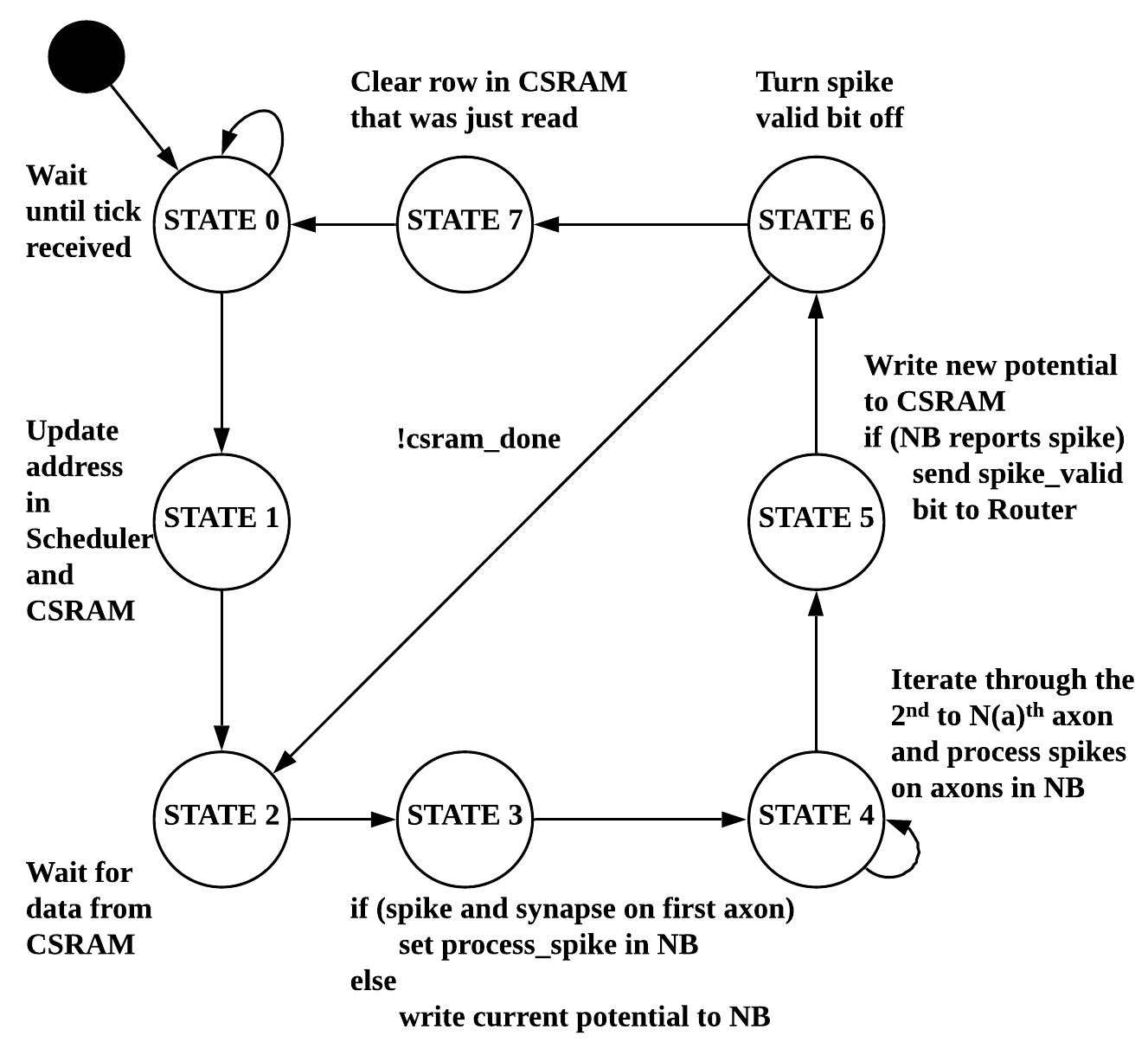}
    \vspace{-3mm}
    \caption{\edit{State machine defining behavior of the \emph{core controller}.}}
    \label{fig:neuron_controller_fsm}
    \vspace{-12pt}
\end{figure}

The \emph{\edit{core controller}}, as shown in Figure~\ref{fig:neuron_controller_fsm}, works with the \emph{neuron block} datapath to coordinate memory accesses and data transfers to ensure that all $N(n)$ neurons inside the core are being emulated faithfully.
For each neuron in the \emph{neuron block}, the \emph{\edit{core controller}} iterates over its associated input axons.
Information regarding the spikes sent to this neuron are retrieved from the \emph{\edit{core SRAM}}, and for each axon that delivers a spike, the \emph{\edit{core controller}} checks if the axon-neuron crossbar contains a connection at this location.
If a connection is present, the weight associated with that axon is sent to the \emph{neuron block} datapath for accumulation into the neuron potential.
Once all input spikes are processed for a given neuron, the controller checks if this neuron has produced an output spike.
If it has, the controller \edit{sends a ``\texttt{spike\_valid}" signal to the router for it to enqueue} this spike for delivery to its destination.
Once all $N(a)\times N(n)$ synaptic connections are processed by the \emph{\edit{core controller}}, it proceeds to an idle state until the next synchronization tick occurs.
If the %
synchronization signal occurs before computation completes (i.e. $f_{tick}$ is too fast), an error flag is raised to notify the user that output may be corrupted.

\subsubsection{\edit{Core SRAM}}

All user-supplied configuration parameters that are relevant for the configuration of each neuron in a single core are stored in the \emph{\edit{core SRAM}}.
It is defined by a matrix of $N(n)$ rows by $B$ columns, where $B$ is the number of bits required to encode all parameters for a single neuron (which varies based on user parameter choices), and there are $N(n)$ data words for all $N(n)$ neurons.
\edit{
All parameters for each neuron (weights, connections, current potential, reset values, thresholds, leak value, destination of generated spikes) are stored as a single word in the core SRAM. 
As such, each neuron requires only one read per tick from the core SRAM. 
Each of these parameters are assigned to their respective inputs in Figure~\ref{fig:neuron_block}, and the neuron processes all N(a) input axons through states 3 to 6 of Figure~\ref{fig:neuron_controller_fsm}. 
After computation is complete, the final $v_j(t)$ potential value is updated and the modified word of the core SRAM is committed back to memory in a single write, after which the core controller proceeds on to the next neuron or to an idle state waiting for the next tick.
}

\subsubsection{Packet Router}

\begingroup
\algrenewcommand{\algorithmiccomment}[1]{\hfill// #1}
\begin{algorithm}[tb]
  \scriptsize
    \caption{Dimension-order/``XY" packet routing}
    \label{alg:router}
    \begin{algorithmic}[1]
        \Function{route}{packet, core, dx, dy}
            \If{dx $<0$}
                \State route(packet, core.east, dx$+1$, dy)
            \ElsIf{dx $>0$}
                \State route(packet, core.west, dx$-1$, dy)
            \ElsIf{dy $<0$} \Comment dx == 0
                \State route(packet, core.south, 0, dy$+1$)
            \ElsIf{dy $>0$}
                \State route(packet, core.north, 0, dy$-1$)
            \Else \Comment dx == 0 \&\& dy == 0
                \State core.accept(packet)
            \EndIf
        \EndFunction
    \end{algorithmic}
\end{algorithm}
\endgroup

\begin{figure}
    \vspace{-4mm}
    \centering
    \includegraphics[width=0.75\linewidth]{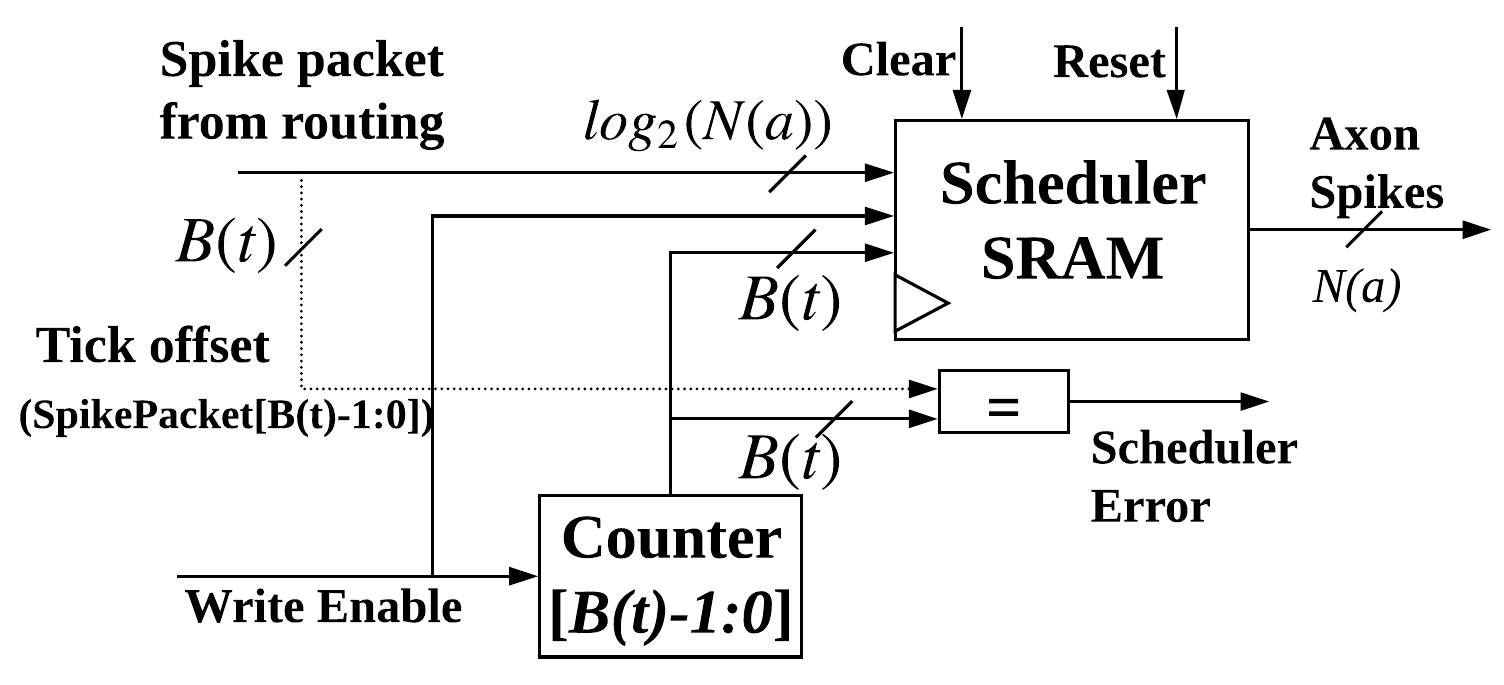}
    \vspace{-2mm}
    \caption{Design of the \emph{packet scheduler} component}
    \label{fig:scheduler}
    \vspace{-12pt}
\end{figure}

With the operation of a single core of neurons defined, the \emph{packet router} ensures that output spikes generated by the neurons in this core can be routed to their destination whether that is back to another input axon in the same core or elsewhere in the NoC. Basic operation is illustrated in Algorithm~\ref{alg:router}.
When a neuron produces a spike, the destination for this spike is retrieved from the \emph{\edit{core SRAM}} and used to construct a \textit{packet} for routing through the mesh.
A packet's destination is encoded as an offset relative to the producer core, and it follows a dimension-order/``XY" routing convention where north and east are positive while west and south are negative.
\edit{
Each packet consists of four key fields: signed relative \textit{dx} and \textit{dy} offsets to route the packet to its destination core, an axon index to identify the destination axon, and a delivery tick offset to identify which tick, relative to the current time, the scheduler should schedule this spike for delivery. 
These last two fields are explored in more detail in discussions in Section~\ref{subsubsec:packet_scheduler}.
}
If, in the course of routing, a destination core is unable to accept an input packet and backpressure must be applied, the router is responsible for further propagating such backpressure to other cores in the system. 
\edit{
Backpressure is applied by giving the receiving core control over enabling reads on the sending core’s FIFO. 
If the receiving core is unable to accept, the sending core’s FIFO will eventually fill, stall, and likewise propagate to its neighbors. 
Each core can route to any other core within the x,y-ranges [-($dim_{x/y}^{max}$)/2, ($dim_{x/y}^{max}$)/2 - 1)].
As such, range is a function of the user-specified $dim_x^{max}$ and $dim_y^{max}$ parameters.
}

\subsubsection{Packet Scheduler} \label{subsubsec:packet_scheduler}

When a packet arrives at its destination, the corresponding core's scheduler \edit{-- shown in Figure~\ref{fig:scheduler} --} is responsible for \textit{scheduling} of this input spike for processing.
An incoming packet is decomposed into two values: the first value is a user-defined offset for how long this input spike should wait before being processed by the core (up to a user-configurable maximum of $N(t)$ ticks), and the second value is a $log_2(N(a))$-bit field that indicates for which axon this spike is intended.
Both values are used to index into an auxiliary memory of size $N(a)\times N(t)$ and write a spike, where the first axis is the destination axon and the second axis is the time offset for arrival.
Concurrently, the scheduler keeps track of the \textit{current} time and, as it cycles through all $N(t)$ possible time slots, it feeds in the input spikes to their destination axons for the core's \emph{neuron block} to process.
If a packet ever attempts to write into the current time offset (i.e. a packet arrives too late), an error flag is thrown and the packet is dropped, but operation otherwise continues unimpeded.

\subsection{Ecosystem Discussion} \label{subsec:ecosystem_discussion}

\begin{figure}[tb]
    \centering
    \includegraphics[width=0.725\linewidth]{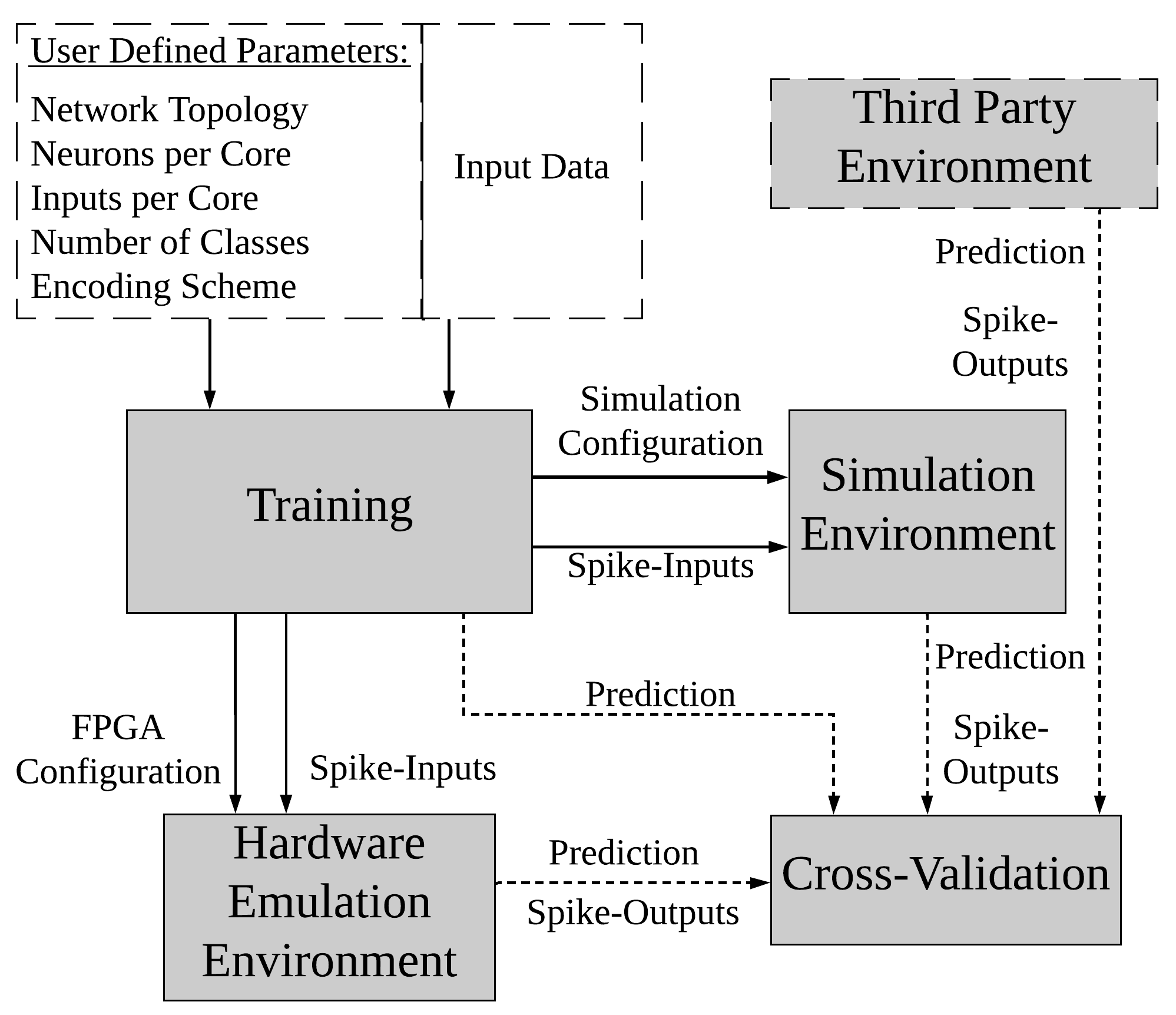}
    \vspace{-2mm}
    \caption{Training, Testing, and Emulation environment}
    \label{fig:full_ecosystem}
    \vspace{-6mm}
\end{figure}

The RANC ecosystem is composed of training, software simulation, and FPGA emulation environments as illustrated in Figure~\ref{fig:full_ecosystem}.

\subsubsection{TensorFlow Training Environment} \label{subsubsec:tensorflow_environment}
The training environment includes a set of libraries built around TensorFlow~\cite{Abadi_2015_Tensorflow} to create RANC-compatible neural networks that target both the simulation and emulation environments.
To map a Spiking Neural Network (SNN) into the RANC environment, we use the methodology from Esser et al.~\cite{NIPS_2015_Training}.
In \edit{the methodology by Esser et al.~\cite{NIPS_2015_Training}}, the layers of the architecture are initialized fully connected, and weights of $\pm 1$ are randomly assigned across all the connections.
Then, during backpropogation, rather than train the weights, \edit{\emph{synaptic connection probabilities}} are maintained as to whether or not a synaptic connection is present between two neurons. \edit{During forward propagation -- either during training or in order to output the final network -- these \emph{synaptic connection probabilities} are converted to either 0 or 1 based on whether the current connection probability is below or above 0.5}
Effectively, the result of this training process is a constrained network containing weights of either $\pm 1$ if a connection is present or $0$ if a connection is not present.
\edit{Connection probabilities are constrained to 0 or 1 in forward propagation throughout training to eliminate quantization error} during final conversion as the network is directly compatible with the RANC hardware throughout the entire training process.
After training is complete, 
the training environment converts the finalized network model and input spike patterns into a format usable by both the software simulation and FPGA emulation environments.
This allows end users 
seamlessly map Tensorflow networks to both the C++ simulation and FPGA emulation RANC environments without necessarily knowing the details of how either work.

\subsubsection{Software Simulation Environment}\label{subsubsec:software_simulation}

The RANC software simulator is a tick-accurate simulation of the \edit{full} RANC architecture that requires two input files.
The first file is a configuration file that specifies global parameters for the simulation such as size and dimensions of the RANC NoC or flags for enabling or disabling debug logging.
The second file specifies the remainder of the configuration required for proper simulator execution.
This includes the input spikes that will be sent into the RANC NoC, the threshold, connection, and reset mode parameters for each neuron, and the core in the NoC from which ``output" will be collected.
After launching the simulator, a trace is collected containing all spikes sent to the ``output core", which can then be analyzed further to determine correctness of an application's behavior.

\subsubsection{FPGA Emulation Environment}\label{subsubsec:fpga_emulation}

To utilize the FPGA emulation environment, we provide an FPGA IP wizard that allows for specifying a set of memory files to configure each of the \emph{\edit{core SRAM}}s present in a given design.
After this, a design is synthesized and contains all information about the connections between neurons as well as their threshold and spiking behaviors.
As such, the IP core can then be deployed into a broader design, with examples using the Xilinx ZCU102 and Xilinx Alveo U250 FPGAs provided in the project documentation.
Spikes are routed into the network via AXI4 communication and results are collected similarly.
Output from the FPGA emulation can be directly compared \edit{tick-by-tick} against the output from the simulation environment, and in the scope of SNNs, \edit{final classification decisions from both} can be validated against output from the Tensorflow environment.

\input{sections_revised/tables/parameterization}

\subsection{\edit{Architectural Extensibility}}

\edit{
One of the primary goals for RANC is ensuring that it is easy to extend in both the software simulation and FPGA emulation environments. 
Each component is modeled independently from the others, and as long as the necessary interfaces are preserved, internal behaviors can be changed without affecting the remainder of the system. 
In the FPGA environment in particular, RANC produces an AXI-compatible IP core that can easily be integrated with other pre/post-processing methodologies. 
If alternative neuron behaviors are desired, as long as the neuron block can receive and output potential values and spikes, the internal behavior can be trivially adjusted. 
If customizations are required in the Router such as using a different NoC topology, this can be accommodated by either changing the logic that modifies the dx and dy packet coordinate bits or through physically changing the for-loops that instantiate and connect the cores.
}
To illustrate the utility of this environment for emulating rich neuromorphic architectures, here we end this discussion by noting that through only \edit{the} choice of configuration parameters, it is possible to behaviorally emulate IBM's well-established TrueNorth architecture within RANC.
TrueNorth cores each consist of 256 input axons and 256 output neurons with four 9-bit weight values per neuron. 
Individual spikes can be routed via relative offset to anywhere within a 256x256 core radius around the spiking core, and their arrivals can be scheduled to occur with up to a 16 tick offset relative to the current tick.
As all of these values are parameters in the RANC hardware design, as shown in Table~\ref{tab:EnvironmentSupport}, we are able to configure RANC to emulate TrueNorth with minimal effort.

As TrueNorth has a rich availability of published literature that details its behavioral and performance characteristics, in Section~\ref{sec:verification}, we perform a number of validation studies between TrueNorth and its emulation in the RANC environment
across traditional neuromorphic applications such as SNN execution and non-traditional applications such as VMM execution.

%% file: sections_revised/tables/glossary.tex
\begingroup
\def\arraystretch{1.3}%
\begin{table}
    \centering
   \caption{Parameters and variables used throughout the RANC ecosystem and architectural diagrams.}
   \vspace{-2mm}
    \begin{tabular}{c|c c}
        \hline
        \edit{Type} & Symbol & Description \\
        \hline\hline
        \multirow{11}{*}{\rotatebox[origin=c]{90}{\begin{tabular}{@{}c@{}}\edit{Neuron} \edit{Param.}\end{tabular}}}
        & $v_{j}(t)$ & voltage potential for neuron $j$ at tick $t$ \\
        & $v_{j}^{+}$ & positive threshold for neuron $j$ \\
        & $v_{j}^{-}$ & negative threshold for neuron $j$ \\
        & \edit{$w_{j}^{k}$} & \edit{weight k for neuron $j$} \\
        & $\ell_{j}$ & leak value for neuron $j$ \\
        & $r_{j}^{+}$ & positive reset value for neuron $j$ \\
        & $r_{j}^{-}$ & negative reset value for neuron $j$ \\
        & $s_{i}(t)$ & spike value on axon i at time $t$ \\
        & $\tau_i$ & axon type for axon $i$ \\
        & $a_i$ & the $i^{th}$ axon in a core \\
        & $n_j$ & the $j^{th}$ neuron in a core \\
        \hline
        \multirow{6}{*}{\rotatebox[origin=c]{90}{\begin{tabular}{@{}c@{}}\edit{NoC} \edit{Param.}\end{tabular}}}
        & $f_{tick}$ & frequency of the global NoC tick \\
        & $f_{core}$ & frequency of a single RANC core \\
        & \edit{$dim_x$} & \edit{X dimension of NoC grid} \\
        & \edit{$dim_y$} & \edit{Y dimension of NoC grid} \\
        & \edit{$dim_x^{max}$} & \edit{Max range of NoC packet in X direction} \\
        & \edit{$dim_y^{max}$} & \edit{Max range of NoC packet in Y direction} \\
        \hline
        \multirow{4}{*}{\rotatebox[origin=c]{90}{\begin{tabular}{@{}c@{}}\edit{Numeric}\\\edit{Param.}\end{tabular}}}
        & $N(a)$ & number of axons per core \\
        & $N(n)$ & number of neurons per core \\
        & $N(t)$ & number of tick slots per core \\
        & $N(w)$ & number of weights supported per neuron \\
        \hline
        \multirow{6}{*}{\rotatebox[origin=c]{90}{\begin{tabular}{@{}c@{}}\edit{Bitwidth}\\ \edit{Param.}\end{tabular}}}
        & $B(a)$ & bits per axon index (i.e. $log_2(N(a))$) \\
        & $B(n)$ & bits per neuron index (i.e. $log_2(N(n))$) \\
        & $B(t)$ & bits to index a single tick (i.e. $log_2(N(t))$) \\
        & $B(w)$ & bits to represent a weight \\
        & $B(v)$ & bits to represent a potential value \\
        & $B(\ell)$ & bits to represent a leak value \\
        \hline\hline
    \end{tabular}
     \label{tab:glossary}
    \vspace{-4mm}
\end{table}
\endgroup

%% file: sections_revised/tables/parameterization.tex
\begin{table}[t]
\caption{
Feature comparison between TrueNorth and RANC. 
}
\vspace{-2mm}
\label{tab:EnvironmentSupport}
\centering
\begin{threeparttable}[b]
\scalebox{1}{
\begin{tabular}{lcc}
    \toprule
                                                         & \multicolumn{1}{|c|}{\textbf{TrueNorth}~\cite{merolla_digital_2011}} &   \multicolumn{1}{c}{\textbf{RANC}} \\
    \midrule
    \multicolumn{3}{l}{\textbf{Basic Functionalities}} \\
    \midrule
    \multicolumn{1}{l|}{Global Synchronous Tick}         & \multicolumn{1}{c|}{1 kHz}          & $f_{tick}$ \\
    \multicolumn{1}{l|}{Axon-Dendrite Crossbar}          & \multicolumn{1}{c|}{$256\times256$} & $N(a)\times N(n)$ \\
    \multicolumn{1}{l|}{Weights per Neuron}              & \multicolumn{1}{c|}{4}              & $N(w)$ \\
    \multicolumn{1}{l|}{Neuron Potential}                & \multicolumn{1}{c|}{9 bit signed}   & $B(v)$ \edit{bit signed} \\
    \multicolumn{1}{l|}{Reset Potential}                 & \multicolumn{1}{c|}{9 bit signed}   & $B(v)$ \edit{bit signed}\\
    \multicolumn{1}{l|}{Weight Value}                    & \multicolumn{1}{c|}{9 bit signed}   & $B(w)$ \edit{bit signed}\\
    \multicolumn{1}{l|}{Leak Value}                      & \multicolumn{1}{c|}{9 bit signed}   & $B(\ell)$ \edit{bit signed}\\
    \multicolumn{1}{l|}{Pos./Neg. Threshold}             & \multicolumn{1}{c|}{9/9 bit signed} & $B(v)$ \edit{bit signed}\\
    \midrule
    \multicolumn{3}{l}{\textbf{Neuron Models}} \\
    \midrule
    \multicolumn{1}{l|}{Leaky-Integrate-and-Fire}        & \multicolumn{1}{c|}{YES}              & YES \\
    \multicolumn{1}{l|}{Linear Reset}                    & \multicolumn{1}{c|}{YES}              & YES \\
    \multicolumn{1}{l|}{Stochastic}                      & \multicolumn{1}{c|}{YES}              & NO  \\
    \midrule
    \multicolumn{3}{l}{\textbf{Advanced Functionality}} \\
    \midrule
    \multicolumn{1}{l|}{Spike/Neuron Processing} & \multicolumn{1}{c|}{Parallel}         & Parallel \\
    \multicolumn{1}{l|}{Synchronicity}                   & 
        \multicolumn{1}{c|}{LAGS\tnote{1}} & LSGS\tnote{2} \\
    \bottomrule
\end{tabular}

}
\begin{tablenotes}
    \item[1] Locally Asynchronous, Globally Synchronous
    \item[2] Locally Synchronous, Globally Synchronous
   \end{tablenotes}
\end{threeparttable}
\vspace{-6mm}
\end{table}

%% file: sections_revised/3_verification.tex
In this section, we evaluate RANC's ability to behaviorally emulate IBM's TrueNorth architecture.
We begin by discussing the Xilinx Alveo-based runtime framework used for conducting each of these experiments.
Then, we recreate two TrueNorth-based case studies described by Yepes et al.~\cite{IJCAI_2017_Training} on the MNIST and EEG datasets.
As these both involve mapping SNNs to RANC, we utilize RANC's Tensorflow integration as shown in Figure~\ref{fig:full_ecosystem} to create networks and map them to both our simulation and emulation environments.
Once the network is trained for a given set of architectural parameters and training data, we output four files: the simulation configuration, FPGA configuration, input spikes, and the Tensorflow test predictions.
The simulation and FPGA configurations contain the information described in Sections~\ref{subsubsec:software_simulation} and~\ref{subsubsec:fpga_emulation}.
Output from both the FPGA emulation and software simulation are then compared against the Tensorflow test predictions and potentially output from other simulation environments -- such as IBM's Compass -- for further verification. 
After this, we conclude by presenting a decisive means of validating our architecture against TrueNorth via tick-by-tick analysis of a deterministic algorithm: VMM.

\subsection{Experimental Setup}

The RANC ecosystem includes a large library of wrapper scripts that help to enable efficient workflows for deploying and testing new architectures.
In this work, we present results only from the Alveo U250 runtime environment.
Experiments with the Alveo were conducted on a system running Ubuntu Server 18.04.4 LTS with Linux kernel 4.15.0-91 and Xilinx Runtime (XRT) version 2.3.1301.
RANC integration with Alveo was performed through use of ``RTL Kernels" in the Xilinx Vitis 2019.2 development environment, and workloads were programmed and dispatched with the XRT OpenCL bindings as shown in examples in the RANC repository.
For neural network workloads, the networks were implemented, trained, and converted for use in the RANC simulation and emulation environments using the automatic conversion tools described in Section~\ref{subsubsec:tensorflow_environment}.

\subsection{MNIST Verification}
\begin{table}[t]
\centering
\caption{Results of the MNIST Experiments}
\vspace{-2mm}
\label{tab:MnistResults}
\begin{tabular}{c|cc|cc}
\toprule
\textbf{}                 & \multicolumn{2}{c|}{\textbf{9 Core Network}}    & \multicolumn{2}{c}{\textbf{30 Core Network}}    \\
\midrule
    \begin{tabular}{@{}c@{}}\textbf{Encoding} \\ \textbf{Window}\end{tabular}
    & \textbf{RANC} & \textbf{\cite{IJCAI_2017_Training}} & \textbf{RANC} & \textbf{\cite{IJCAI_2017_Training}} \\
\midrule
1        & 96.65              & 93.73 $\pm$ 0.21               & 97.45              & 95.36 $\pm$ 0.17               \\
2        & 97.19              & 96.07 $\pm$ 0.20               & 97.99              & 96.45 $\pm$ 0.12               \\
4        & 97.40              & 97.28 $\pm$ 0.11               & 98.26              & 97.38 $\pm$ 0.10               \\
8        & 97.06              & 97.75 $\pm$ 0.06               & 98.11              & 97.70 $\pm$ 0.04               \\
16       & 97.55              & 97.95 $\pm$ 0.08               & 98.14              & 97.88 $\pm$ 0.05               \\
\bottomrule
\end{tabular}
\vspace{-5mm}
\end{table}

Starting with the MNIST case study, the architectures used are equivalent to the 9 and 30 core networks presented in Yepes et al.~\cite{IJCAI_2017_Training}.
In these experiments, $N(n) = N(a) = 256$ to enable comparison with TrueNorth-based results. 
\edit{As discussed in Yepes et al.~\cite{IJCAI_2017_Training}}, MNIST images are $28\times 28$, \edit{and this} exceeds the maximum input size of a single core, \edit{so they partition the} input into windows of either 16$\times$16$=$256 or 8$\times$8$=$64 and \edit{feed it} into multiple cores.
The input images are encoded via burst encoding over an input window ranging from 1 to 16 ticks.
Training is performed with dropout between the layers and \edit{the optimizer is not specified.
As such, we choose to utilize the Adam optimizer~\cite{Kingma_2015_ADAM}.
}
No data normalization is applied. 
We present our comparison with the results of Yepes et al.~\cite{IJCAI_2017_Training} in Table~\ref{tab:MnistResults}, and we find that, across the board, we have comparable performance to published work.
We attribute slight deviations in network accuracy to our particular choice of the Adam optimizer and the stochastic nature of neural network training.

\subsection{EEG Verification}

\begin{table}[t]
  \centering
  \caption{Results of the EEG Experiments}
  \vspace{-2mm}
  \label{tab:EEGResults}
  \begin{tabular}{c|c|cc}
  \toprule
  \textbf{Encoding Window} & \textbf{RANC} & \textbf{\cite{IJCAI_2017_Training}} \\
  \midrule
  1                        & 67.70              & 62.36 $\pm$ 5.27               \\
  2                        & 62.11              & 66.96 $\pm$ 4.22               \\
  4                        & 65.84              & 70.06 $\pm$ 3.98               \\
  8                        & 75.16              & 72.36 $\pm$ 2.92               \\
  16                       & 75.78              & 75.96 $\pm$ 2.17               \\
  \bottomrule
  \end{tabular}
  \vspace{-5mm}
\end{table}

After MNIST, we verified our environment against the EEG dataset results also presented by Yepes et al.~\cite{IJCAI_2017_Training} using an identical network topology.
The training and testing methodologies applied are also equivalent to the MNIST verification above, and the results are displayed in Table~\ref{tab:EEGResults}.
We find that for all burst-encoding windows tested, our networks achieve quite comparable accuracy with respect to the published results.
\edit{Large variability in achieved accuracy can be attributed to the limited size of the EEG dataset (1109 usable samples) coupled with variations in train-test dataset splitting.}
The outcomes of these experiments give us confidence in our architecture's ability to replicate the commercial TrueNorth architecture, but decisive validation, without having access to the exact network model parameters used in these experiments, is impossible.
As such, we conclude our architectural verification by mapping VMM, a known deterministic application, whose output can be reliably compared tick-by-tick with the source architecture.

\subsection{VMM Verification}

To present a deterministic verification of our architecture's ability to replicate TrueNorth, we implement signed VMM in RANC by using Fair's method of mapping VMM on TrueNorth~\cite{fair19}.

\edit{VMM is an ideal application for validating our emulation environment as each problem instance has a unique, deterministic output, and its implementation on TrueNorth requires spreading computation across a network of interconnected cores and running for hundreds of ticks. To produce equivalent behavior, each of these interconnected RANC cores must produce the same output spikes on the same tick as TrueNorth.}

\edit{We use a 9-bit signed VMM formulation, and validate the behavioral functionality of both our emulation and simulation environments against an equivalent implementation in IBM’s Compass [29] environment. We created 100 random matrix-vector pairs with the matrices ranging from 2x3 to 8x8 and the vectors sized appropriately for their corresponding matrix consisting of random 9-bit signed integers. Each of these matrix-vector pairs was mapped to Compass, the RANC simulator, and the RANC emulator using the method proposed by Fair et al. [26]. Across all instances, we found a one-to-one match not only in the decoded VMM output but also in tick-level timing of when various outputs were observed.} In conclusion, we are able to emulate TrueNorth in the RANC ecosystem using traditional SNN applications (MNIST, EEG) and non-traditional deterministic application (VMM).
In the following section we present detailed case studies on RANC's ability to drive neuromorphic architecture exploration through VMM and \edit{convolution} based classification. 

%% file: sections_revised/4_case_studies.tex
With the functional validation of our architecture complete, we exploit the modularity of the RANC ecosystem by exploring methods through which an architect or application engineer can incrementally change the behavior of the hardware based on application bottlenecks -- a task infeasible for researchers using either TrueNorth or Loihi due to the fixed nature of the chips.
We begin with an in depth analysis of mapping VMM onto a neuromorphic architecture, and we then demonstrate VMM execution by conducting a tick-by-tick analysis and visually monitoring the state of key components. This detailed analysis allows us not only to understand the interactions between processing and routing elements essential for realizing neuromorphic behavior but also expose optimization opportunities.
We then expand our architectural optimization efforts 
\edit{towards exploring the use of customized convolutional cores in improving neuron utilization, axon utilization, and hardware throughput via case studies on synthetic aperture radar (SAR) and CIFAR-10 image classification.}
By again exploiting the ability to reconfigure components for each application, we demonstrate that a modified RANC architecture vastly reduces resources required for mapping convolutional network layers by allowing each core to be fully utilized.
Finally, we conclude our case studies with FPGA based measurements on resource usage, scalability, and path delay for a variety of architectural configurations.

\subsection{VMM Optimization}\label{subsec:vmm}
\edit{In the following subsection, we illustrate the resource utilization inefficiencies that are present when dealing with signed VMM execution on TrueNorth, and we demonstrate how RANC's \textit{neuron \edit{datapath}} resolves this inefficiency.}
\edit{For interested or unfamiliar readers, a preliminary background on the non-signed, positive-only VMM mapping approach taken by Fair et al.~\cite{fair19} is provided in Appendix~\ref{app:vmm}.}

\subsubsection{Complications of signed VMM}
As the rate encoded input spikes lack sign, to make signed VMM mappable to the TrueNorth, Fair et al.~\cite{fair19} %
\edit{divides axons and neurons} into positive and negative groups, where positive and negative input spikes are routed to their respective groups
allowing each group to represent positive and negative outputs from respective connected axons. 
\begin{figure}[t]
    \centering
    \includegraphics[width=0.75\linewidth]{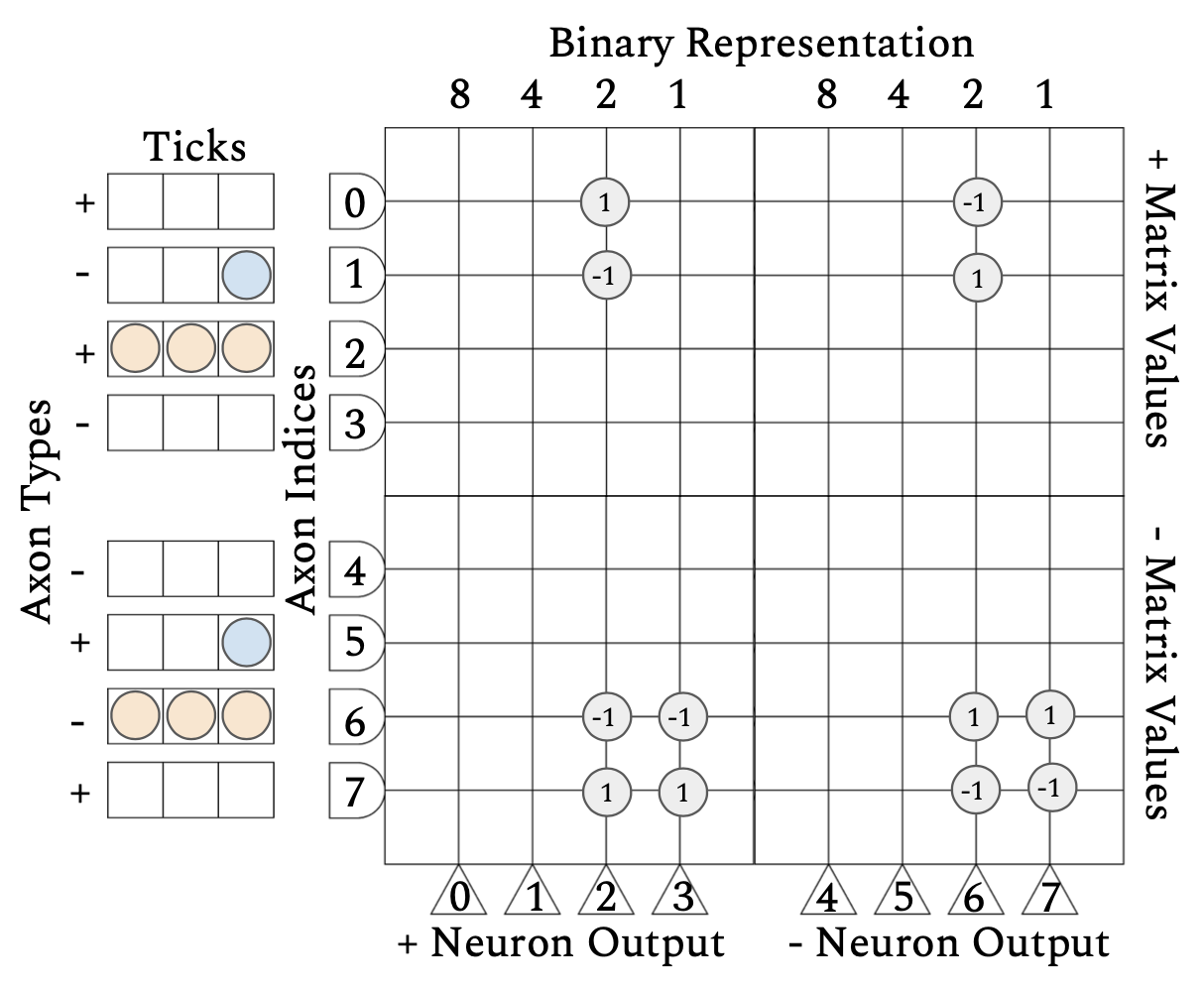}
    \vspace{-2mm}
    \caption{Mapping scheme for first core of +/- VMM. Axons are divided into alternating +/- representation for signed inputs , and subdivided into upper(+) and lower(-) groups for signed matrix values, with + values occupying upper 4 rows, and - lower 4 rows. %
    }
    \label{fig:pos-neg-mapping}
    \vspace{-5mm}
\end{figure}
We illustrate Fair's representation using Figure~\ref{fig:pos-neg-mapping} based on an example multiplication of input vector [-1 3] with the input matrix [2 -3]$^T$. The single crossbar with 2 axons and 2 neurons is replaced with four crossbars in order to cover all possible combinations of  positive and negative representations of axons and neurons. 

To map this approach to TrueNorth, additional changes are needed, however.
As the neuron datapath in TrueNorth uses a $\geq$ comparison in positive threshold comparison but a $<$ comparison on the negative threshold, Fair et al.~\cite{fair19} has shown that additional feedback systems are needed to correct behaviors associated with this asymmetric thresholding.
Uncorrected, this asymmetry allows the potential value of a neuron to remain negative when it should be otherwise reset to zero.
Correct output can only be achieved with additional feedback that reroutes spikes back to the core and drives the potential back to zero.

\begin{figure}[t]
    \centering
    \includegraphics[width=0.85\linewidth]{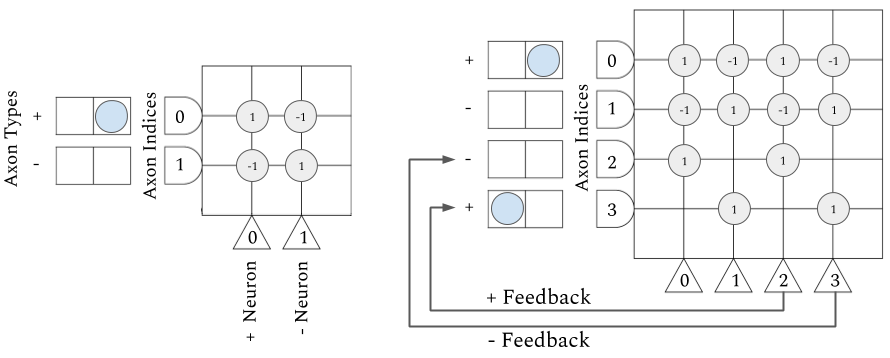}
    \vspace{-2mm}
    \caption{RANC allows signed VMM without feedback (left) while the equivalent operation on TrueNorth requires feedback (right).}
    \label{fig:feedback}
    \vspace{-6mm}
\end{figure}

This feedback concept is realized by duplication of the neurons, and grouping them into positive and negative feedback neurons. We show this duplication in Figure~\ref{fig:feedback}, where feedback neurons route spikes back to a newly created group of axons within the same core a tick later.
These new axons are divided into groups for their connected positive and negative feedback neurons.

In Figure \ref{fig:feedback}, \edit{Axons 0 and 1} ($a_0$ and $a_1$) are the positive and negative input axons, and \edit{Neurons 0 and 1} ($n_0$ and $n_1$) are positive and negative neurons respectively. Feedback is connected by duplicating the $n_0$ and $n_1$ to $n_2$ and $n_3$ and routing them back to $a_2$ and $a_3$ as positive and negative feedback channels.
The connections aligned with $n_2$ and $n_3$ are feedback specific connections, which ensure feedback spikes connect to neurons that have neuron potential of -1. To illustrate this, a spike is arbitrarily placed on $a_0$. During the first tick, the spike produces neuron potentials of [1, -1, 1, -1] for each of the neurons [$n_0$, $n_1$, $n_2$, $n_3$]. The asymmetric thresholds restrict the reset method from bringing the negative potentials back to zero, producing a next tick potential of [0, -1, 0, -1] for the neurons. Feedback corrects asymmetric thresholds through spike loopback. As $n_2$ had a neuron potential $\geq$ 1, a spike is emitted and redirected back at $n_3$. During the next tick this spike connects at $n_1$ and $n_3$ producing neuron potential of [0, 0, 0, 0]. This works exactly the same if the initial spike were to occur on $a_1$ instead of $a_0$. Additionally, this functions just as well should there be another spike behind the one originally set on $a_0$, due to the additive effect of spikes, it balances out by canceling the negative neuron potential.

The neuron duplication and feedback based workaround due to architectural constraints is resolved in RANC with a neuron datapath that supports symmetric threshold behavior. Replacing $<$ with $\leq$ for the negative threshold comparison allows negative potentials to reach zero without having to use feedback, ensuring that a positive input spike on these neurons is correctly counted. 

Fair’s 8x8 VMM implementation is a three core setup on TrueNorth, where the first core  receives the input spikes representative of the vector by which to multiply. 
The output of the first core is an unsigned binary representation of the output vector. 
Due to the replication overhead for feedback as illustrated for the 2x2 example in Figure~\ref{fig:pos-neg-mapping}, this core requires 160 axons and 256 neurons for the 8x8 VMM implementation. The second core contains neurons with synaptic weights of \{8, 4, 2, 1\}, which applies weighting to the binary representation and requires 128 axons and 32 neurons.
The final core, which is a {16, 1} weight application, applies significance to the four most- and four least-significant bits, as well as performs the summation in order to generate the final output of the VMM operation. The third core requires 48 axons and 32 neurons. 
Therefore, the 8x8 VMM implementation requires a total of 336 axons and 320 neurons. 
Eliminating the feedback system through symmetric thresholds, leaves behind a 32 axon, 128 neuron setup for the first core of the VMM implementation as the 128 feedback neurons and axons can be discarded. Similarly, the subsequent two cores (magnitude, weight application) require 128x32 and 32x16 axons and neurons respectively. Therefore VMM mapping on RANC requires a total of 192 axons and 176 neurons, reducing the axon and neuron footprint by 57\% and 55\% respectively.  

\begin{table}[t]
\centering
\tabcolsep=0.12cm
\caption{
FPGA resource utilization on Alveo U250 for 9-bit signed VMM using three cores. Each row represents a core tested with \edit{RANC-emulated} IBM TrueNorth (TN) with feedback, and RANC with symmetric threshold. Second core requires no feedback, and is presented for completeness. }
\label{tab:fpga}
\begin{tabular}{|c|ccccc|}
\hline
Core           &  LUT & RAM$_{LUT}$ & FF   & BRAM  & \begin{tabular}[c]{@{}l@{}}Delay (ns) \end{tabular} \\ \hline
$TN^{Core1}$            &  1500    & 192    & 1076 & 4 & 3.925                                                \\
$RANC^{Core1}$            &  1038    & 48     & 937  & 2 & 3.801                                                \\ \hline
Core2            & 1559    & 266    & 1244 & 0 & 3.928                                                 \\ \hline
$TN^{Core3}$            & 1200    & 130    & 1112 & 0 & 3.771                                                 \\
$RANC^{Core3}$           & 1106    & 105    & 1044 & 0 & 3.563                                                 \\ \hline
Savings(\%) &           20.6 & 52.5   & 9.5  & 50.0 & 2.9                                                \\ \hline
\end{tabular}
\vspace{-12pt}
\end{table}

In Table~\ref{tab:fpga}, we compare the FPGA resource utilization of functionally equivalent signed VMM implementations on the emulated reference TrueNorth and default RANC configuration across cores one and three. Core two is excluded since number of axons and neurons required by this core are same for both architectures. We use the $1\times8$ vector and $8\times8$ matrix as this VMM problem occupies an entire core of TrueNorth. We observe the highest saving in the first core, where signed operations are carried out with reduction of 30.8\% and 75\% for LUTs and LUTRAM respectively. 
In the third core, we observe relatively less savings in LUT (7.8\%) and LUTRAM (19.2\%) usage compared to the first core as the number of axons and neurons removed is much fewer. 
Overall, the 55\% reduction in neuron count corresponds to reduction in BRAM and LUTRAM usage by 50\% and 52.5\% respectively. 
As we reduce the neuron footprint of the VMM mapping, muxes, adders and comparators contribute to reduced LUT usage and the neuron potential register contribute to the LUTRAM reduction. Without feedback neurons, the 128 axons to which they were connected are discarded. This decreases size of the \emph{packet scheduler} that is entirely mapped onto LUTRAM. 
Additionally, removing the feedback reduces total latency slightly from 11.624ns to 11.292ns (2.9\%).

Symmetric behavior of neuron potential thresholds in RANC eliminates the need for resource duplication based workaround and allows for improved scalability to process larger matrix sizes and implement applications that involve VMM operations such as convolution, locally competitive algorithm, or least squares minimization.
In the next case study we 
demonstrate our ability to customize RANC for the granularity of convolution operations. 

\subsection{\edit{Convolution}}
\label{subsec:sar}

Mapping convolution onto neuromorphic architectures is either restricted in size or is often inefficient due to the need for extensive repetition of input data \cite{PNAS_2016_Applications}. From size point of view, for instance, the best CIFAR10 network reported by~\cite{PNAS_2016_Applications} requires 31,492 cores. Given that CIFAR images are only 32x32, there are scalability challenges for applications that require processing larger images such as 512x512 images from the EM segmentation challenge dataset~\cite{arganda2015Crowdsourcing}.From inefficiency point of view, studies have shown that crossbar underutilization is a challenge when mapping neural network connections to neuromorphic architectures \cite{song2020compiling, Yi18}.

The inefficiency of mapping convolution can be reduced by increasing the percent of the total image available within a single core. 
If the entirety of the image can be sent to a core, then the need for all axon redundancy is eliminated and a larger fraction of the core's neurons can be used. 
This is because all possible strides \edit{of each convolutional filter} can be calculated as every input is available for computation. 
\edit{When an image can not fit entirely within a single core, computations must be split across multiple cores, and pixels that are shared between kernels located in different cores must be replicated.} 
An alternative method for addressing the mapping inefficiency for convolution could be scaling down the kernel size for a layer, or increasing feature counts. This approach allows more strides to be calculated within a single core and decreases the amount of input data replication required for the next core. However, network design decisions should not be forced upon the designer \edit{purely} due to \edit{restrictions} of the deployment hardware. \edit{Instead, they should be guided collectively by insights from both hardware and software design perspectives to jointly optimize application performance and hardware efficiency, and they should converge to values that enable the application designer to maximize performance without requiring expensive tradeoffs caused by mismatch in hardware capabilities.}

\begin{figure}[t]
    \centering
	\includegraphics[width=0.95\linewidth, %
	]{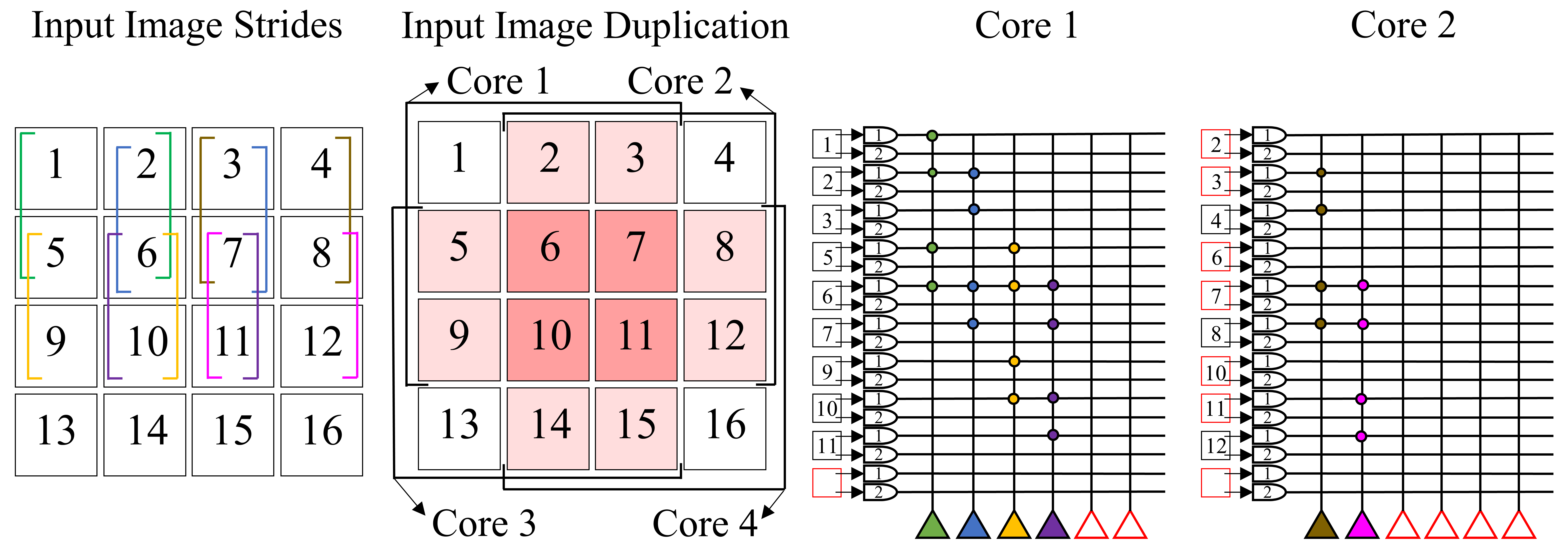}
 	\caption{\edit{Example convolution mapping  with 4x4 image, 2x2 kernel and weights of all 1. Kernel strides mapped to a core are indicated by a matching color. Core 1 calculates output for 4 pixel sets, while cores 2-4 calculate the remaining 5. The color red denotes where axon duplication or neuron waste occurs. Core 2 receives 6 pixels (2,3,6,7,10,11) that are already provided to Core 1. Core 3 contributes with only one unique convolution (11,12,15,16), where its remaining three convolution operations are covered by the other cores.}}
    \label{fig:conv}
    \vspace{-12pt}
\end{figure}

\edit{Figure~~\ref{fig:conv} provides an illustration of the convolution mapping~\cite{PNAS_2016_Applications} that utilizes a neuron to represent a single output feature of a convolutional stride, with each connected axon supplying an input value from the image. 
In this example, we assume that the image is 4x4, the kernel is 2x2, and each core has 20 axons with 6 neurons. 
Two axons are used per input pixel to allow for ternary kernel weights of -1, 0, or 1. 
Ternary weights limit this axon configuration to process a maximum of 10 pixels worth of input data per core, therefore a core can process a maximum of 3x3 input region that allows for a maximum of 4 cores. 
Due to the overlapping regions, only 9 of the 24 total neurons are utilized in overall.
Pixels 6, 7, 10, and 11 are replicated in all 4 cores. 
Pixels within the middle of the perimeter of the image are replicated twice. 
Among all 4 cores, unique axon usage is 40\%, and 60\% of axons are either unconnected or receiving a duplicate input. }

\edit{Adjusting the core input size to accommodate the full image} is infeasible within fixed neuromorphic architectures as \edit{the per-core} axon count is a set value across the system, typically well below the number required for an entire image. The RANC emulator allows the designer to configure axon and neuron counts per core, which can be tuned to be a function of the network's input image size and kernel parameters.
\edit{For example, in the provided example above, a custom convolutional core of 32 axons and 9 neurons perfectly maps the operation with no waste. Although the individual core resource usage itself is greater, the same operation that originally required 4 cores or 80 axons now uses a single core with 32 axons. 
\edit{While this larger core would come at the cost of a slower tick rate as the computations required within the core controller increase, this tradeoff can be seen as desirable by reducing resources enough such that many parallel classifiers can be instantiated with the savings. To explore this, in this section, w}e utilize RANC's ability to deploy heterogeneous cores and propose dedicated convolution cores that reduce the need for input redundancy. 
By increasing the axon and neuron counts per core, we show that we are able to improve axon utilization without sacrificing the classification accuracy. 
For this, we set up a controlled experiment starting with the RANC architecture defined based on the VMM study that supports symmetric threshold. We refer to this implementation as the \emph{Default} configuration. 
\edit{We establish a common data preparation and network training flow that is shared between the two configurations, and as such the only variation occurs on the architecture.}
\edit{By controlling variation to only be within the architecture, we evaluate the impact of the proposed heterogeneous core configurations on hardware performance with respect to post-routing FPGA resource usage, latency, and classification throughput with networks trained for Synthetic Aperture Radar (SAR) and CIFAR-10 image classification. We end by noting that, as these changes affect only architectural mapping strategies, the network behavior is unchanged, and as such, classification accuracy is not compromised.}
}

\subsubsection{SAR Dataset, Training and Mapping}
The MSTAR public dataset \cite{MSTAR} is a collection of SAR images 
that contains eight classes of military vehicles, in which the target is centered and taken from an angle of depression of 15 and 17 degrees. Images are grayscale and roughly 128x128 in dimension, and each testing class contains around 300 images. 
\begin{figure}[t]
    \centering
    \includegraphics[width=0.85\linewidth]{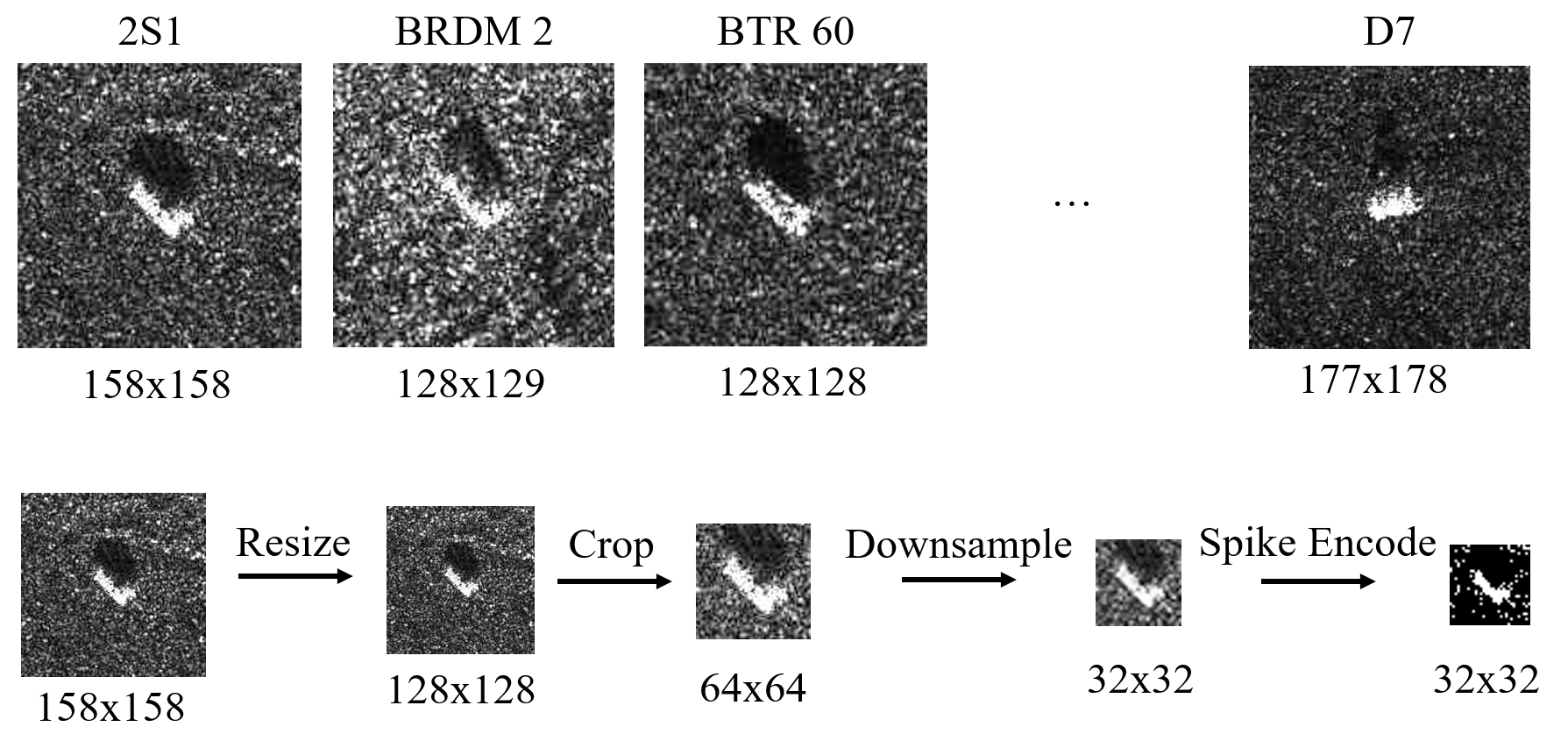}
    \vspace{-2mm}
    \caption{MSTAR samples and RANC's preprocessing based on~\cite{Renz17}}
    \label{fig:MSTAR_classes}
    \vspace{-6mm}
\end{figure}
To map SAR classification to the RANC ecosystem, we first preprocess the imagery in a similar approach taken by Renz and Wu~\cite{Renz17} as illustrated in Figure~\ref{fig:MSTAR_classes}. MSTAR dataset images vary in size from class to class, so 
we normalize all images to 128x128 first, then take a 64x64 image chip from the center of this resized image. These 64x64 images are then down-sampled to 32x32 using nearest-neighbor sampling. 
The grayscale pixels, which take values between 0 and 255, are encoded into a spike form via a simple thresholding method where values $<$128 are dropped and values $\geq$128 are converted \edit{to a single input spike}.
The converted spikes are then sent into our network as a representation of the target object.
The network we deploy uses a convolutional layer followed by two fully connected layers. The result is then passed through an additive pooling layer for final classification to one of the eight classes. 

We ran sweeping experiments with RANC's training libraries to determine the ideal kernel and stride sizes within the ranges of 4 to 24 and 1 to 5 respectively. Based on these experiments, our implementation's convolutional layer uses two 11x11 kernel feature maps with a 1x1 stride, which generates an output feature count close to the 1024 inputs available into next fully connected layer. The fully connected layers are trained using the flow presented in Section~\ref{sec:ecosystem}. 
\edit{
To build SNN-compatible 2D convolutional layers, we train standard 2D convolutional layers and quantize them. 
Quantization from floating-point is done after training is complete to create an SNN-compatible layer, and these are linked with the fully connected layers trained with the methodology from Section~\ref{sec:ecosystem}.
}
We then feed these trained weights along with the spike encoded MSTAR dataset into the RANC ecosystem for our functional verification and accuracy analysis. 
\edit{While better training techniques such as BinaryConnect~\cite{courbariaux_binaryconnect_2015} exist in the literature for achieving peak accuracy in mapping CNNs to neuromorphic hardware, in this study we have a different objective than pure training accuracy. 
Instead, we seek to illustrate, holding the training method fixed, that there is potential for hardware architectural exploration in optimizing the mapping of existing CNN network structures to neuromorphic hardware platforms.}

\subsubsection{Convolution Mapping Efficiency}

\edit{
In RANC's default configuration, core axon size limits the maximum allowed kernel and image subsection available to a core to be an 11x11 square region. Following the same calculations within the mapping methodology of Figure~\ref{fig:conv}, we find that 484 cores are needed and each pixel is replicated on average 60 times.
By utilizing a modified architecture with heterogeneously sized cores of size 1024x256, we find that a 22x22 image subsection is available to each core, and only four convolutional cores are necessary to compute all overlap regions.
In this methodology, neuron utilization increases from 0.7\% to 94.5\% and effective axon utilization increases from 1.7\% to 50\%.
Details on these calculations are provided in Appendix~\ref{app:sar}.
}
The convolution outputs for both the default and proposed architecture are sent into a fully connected layer of four cores, and each feed 64 neuron outputs to a single core. The output of this last core's 64 neurons can then be classified after using additive pooling to sort them into one of the eight classes.

\subsubsection{Performance Analysis - Default vs. Modified RANC Configurations}

\begin{table}[t]
\centering
\caption{Resources and Performance Comparison for SAR Classification Networks}
\vspace{-2mm}
\label{tab:SARmapping}
\begin{tabular}{|c|cccc|}
\hline
Design         & Cores & Axons  & Neurons & Accuracy \\ \hline
Default        & 489   & 125184 & 125184  & 75.7     \\
Modified       & 9     & 5376   & 1344    & 75.7     \\
Reduction (\%) & -     & 95.7   & 98.9    & 0.0      \\ \hline
\end{tabular}
\vspace{-5mm}
\end{table}

\begin{table}[t]
\centering
\caption{SAR Post-Implementations using Alveo U250 FPGA based on Default, Modified and 24 parallel instances of Modified Configurations with resource usage and ratio with respect to total available for each resource type.}
\label{tab:SARresource}
\begin{tabular}{|c|cccc|}
\hline
Design                & LUT     & RAM$_{LUT}$ & FF      & BRAM   \\ \hline
Default               & 1334047 & 148496 & 751354  & 2684   \\
Default(\%)           & 77.2    & 18.8   & 21.7    & 99.9   \\ \hline
Modified              & 52609   & 7476   & 29754   & 64.0   \\
Modified(\%)          & 3.0     & 0.9    & 0.9     & 2.4    \\ \hline
Modified$^{24}$       & 879965  & 152387 & 386714  & 2152.0 \\
Modified$^{24}$(\%)   & 50.9    & 19.3   & 11.1    & 80.1   \\ \hline

\end{tabular}
\vspace{-5mm}
\end{table}

\begin{table}[t]
\caption{Delay and Throughput Comparison Post-Implementation for SAR Network on Alveo U250}
\vspace{-2mm}
\label{tab:SARtiming}

\begin{tabular}{|c|cccc|}
\hline
Design        & Delay  & Freq       & Clock Cycles & Throughput   \\
              & (ns)   & (MHz)      & per Tick     & (Images/sec) \\ \hline
Default       & 4.688  & 213.3      & 66308        & 3216         \\
Modified      & 4.024  & 248.5      & 265220       & 936          \\
Modified x24  & 4.352  & 229.7      & 265220       & 20785        \\ \hline
\end{tabular}
\vspace{-5mm}
\end{table}

\begin{table}[t]
\centering
\caption{\edit{CIFAR-10 Post-Implementations using Alveo U250 FPGA based on Default, Modified and 8 parallel instances of Modified Configurations with resource usage and ratio with respect to total available for each resource type.}}
\label{tab:CIFARresource}
\begin{tabular}{|c|cccc|}
\hline
Design                & LUT     & RAM$_{LUT}$ & FF    & BRAM       \\ \hline
Default               & 1065424 & 112624      & 595321  & 2035.0   \\
Default(\%)           & 61.7    & 14.2        & 17.2    & 75.7     \\ \hline
Modified              & 112587  & 22094       & 50843   & 245.5    \\
Modified(\%)          & 6.5     & 2.8         & 1.5     & 9.1      \\ \hline
Modified$^{8}$        & 847652  & 169939      & 333971  & 2301.5   \\
Modified$^{8}$(\%)    & 49.1    & 21.5        & 9.7     & 85.6     \\ \hline

\end{tabular}
\vspace{-5mm}
\end{table}

\begin{table}[t]
\caption{\edit{Delay and Throughput Comparison Post-Implementation for CIFAR-10 Network on Alveo U250}}
\vspace{-2mm}
\label{tab:CIFARtiming}

\begin{tabular}{|c|cccc|}
\hline
Design        & Delay  & Freq  & Clock Cycles & Throughput   \\
              & (ns)   & (MHz) & per Tick     & (Images/sec) \\ \hline
Default       & 3.979  & 251.3 & 66308        & 3788         \\
Modified      & 3.864  & 258.8 & 134148       & 1929         \\
Modified x8   & 4.288  & 233.2 & 134148       & 13907        \\ \hline
\end{tabular}
\vspace{-5mm}
\end{table}

In Table \ref{tab:SARmapping}, we compare the resources required by the default architecture and our modified architecture with heterogeneous cores on RANC in terms of number of cores, axons, neurons, and the accuracy of the classification. Although core sizes differ preventing a fair comparison in core count, the heterogeneously sized core requirements see significant reduction in total system axon and neuron count. 
The convolutional layer itself remains functionally equivalent between the two architectures. As only the mapping methodology is changed, no difference in classification accuracy occurs when using the same set of trained weights on the RANC simulator.

In Table \ref{tab:SARresource}, we present post-implementation resource utilization for both architectures on Alveo U250 FPGA. We observe resource savings with the modified architecture
due to factors related to network mapping efficiency. 
Replacing the fully connected 256x256 cores with cores sized at 256x64 within the modified design allows those cores' neuron parameters to map to LUTRAM instead of BRAM, reducing total BRAM usage. 
As in the prior VMM experiment, reducing the neuron count for these cores requires fewer muxes, adders and comparators decreasing LUT usage, and fewer neuron potential registers contribute to LUTRAM reduction. Additionally, a reduction of total core count due to the use of convolutional cores results in large BRAM savings over the default design, as less core parameters will need to be mapped in total.

The reduction in LUTs and BRAMs has the direct effect of improving critical path delay as shown in Table~\ref{tab:SARtiming}. After converting these delays to a clock frequency for both designs, we see that the modified network can operate at a slightly faster clock rate of the RANC default. However, this improvement does not translate to an increased image classification throughput. As largest core size increases within a design, the clock cycles required to complete the \emph{\edit{core controller}'s} state machine increases. The increase is almost directly proportional to the size increase of the core's neuron crossbar. A 1024x256 core's maximum number of synaptic connections is four times of the default 256x256 core, resulting in a tick rate approximately a fourth of the design with default configuration. Coupled with the operational frequency, throughput of the modified design is 29.1\% of the default.

Although the time required for core computations within each tick increases by a factor of four within our modified design, a 97.6\% reduction in BRAM usage allows 24 of these total modified networks to be mapped in parallel before the BRAM usage in the U250 surpasses 80\% as shown in Table \ref{tab:SARresource}. We limit the parallelization past 80\% BRAM usage to achieve a similar operating frequency of the default network. With 24 classifiers in parallel, the heterogeneous core implementation can achieve an image classification throughput 6.46 times that of the default configuration when factoring in the slower tick rate. Note that the default architecture can not be parallelized by any factor due to its large size, and that the 24 modified networks enjoy less or equivalent total usage compared to the default for all resource types. The throughput comparison for each configuration is detailed in Table \ref{tab:SARtiming}.

Renz and Wu deployed a DCNN on IBM's TrueNorth to classify SAR images and achieved an accuracy of 95.6\% \cite{Renz17} with the expense of utilizing 4042 of the 4096 cores on a single TrueNorth chip. Because of the insufficient information on their network architecture and unavailability of IBM's Energy-Efficient Deep Neuromorphic Network (EEDN) learning framework \cite{SC_2016_Foundation}, we were unable to create the exact architecture. Even with Renz and Wu's network design, EEDN does not grant a user access to layer specifics such as regularization or activation functions \cite{DATE_2017_Training}, which would prevent us from accurately replicating their work. However, with limited information and a fewer number of cores, we are able to achieve an accuracy of 75.7\%.
Despite our lower accuracy, the objective of the SAR mapping experiment is to demonstrate the versatility of our hardware and software codesign environment. 
The ability to customize the hardware configuration by adjusting the number of axons and neurons per core allows resource efficient implementation of SAR classification on a neuromorphic architecture. The maximum kernel size limitation for neuromorphic convolution is removed as the number of axons can be flexible per core. These two benefits allow the convolution operation to be mapped more efficiently within deep CNNs, increasing throughput via the ability to parallelize with freed resources.

\subsubsection{\edit{CIFAR-10}}
\edit{
To demonstrate the generalizability of the techniques illustrated with SAR, we repeated the study with a network trained for CIFAR-10.
With an additional two channels of data to represent full RGB, maximum kernel size was further restricted to 7x6 within RANC's 256 axon default configuration. 
The largest initial network we could fit used a convolutional kernel of 6x6 requiring 364 cores, and after applying the same core-sizing optimizations as in SAR, with the modified configuration we were able to reduce to 23 cores.
This corresponds to an 85\% reduction in BRAM utilization as shown in Table~\ref{tab:CIFARresource}.
Similar to the SAR study, the modified network results with an increase in the number of clock cycles per tick due to increase in the size of the neuron crossbar as shown in Table~\ref{tab:CIFARtiming}. 
However, the reduction in resource usage allows achieving a net throughput improvement of 3.6x compared to the default configuration by scaling the modified architecture to eight parallel classifiers. 
Hence, regardless of the dataset type and shape, these techniques can be applied to optimize the resource utilization of CNNs with large amounts of overlap to neuromorphic hardware through architectural adjustments in RANC.
}

\subsection{Scalability Analysis}\label{subsec:alveo_scalability}

\begin{table}[t]
    \centering
    \tabcolsep=0.16cm
    \caption{Alveo U250 Resource utilization for RANC grids composed of $N(a) = N(n) = 256$ cores as a function of grid size}
    \vspace{-2mm}
    \label{tab:alveo_scalability}
    \begin{tabular}{|c|ccccc|}
        \hline
        Cores        & LUT     & RAM$_{LUT}$ & FF & BRAM   & Delay$_{ns}$ \\ 
        \hline
        1$\times$1   & 23152   & 448    & 16777   & 5.5    & 3.835     \\
        2$\times$2   & 27458   & 1056   & 20203   & 16.5   & 3.890     \\
        4$\times$4   & 55229   & 4704   & 38926   & 82.5   & 3.857     \\
        8$\times$8   & 166646  & 19296  & 114142  & 346.5  & 3.821     \\
        16$\times$16 & 616965  & 77664  & 419326  & 1402.5 & 3.962     \\
        20$\times$20 & 1168845 & 121440 & 649815  & 2194.5 & 4.102     \\
        21$\times$21 & 1287726 & 133904 & 715367  & 2420.0 & 4.130     \\
        22$\times$22 & 1412498 & 146976 & 784270  & 2656.5 & 4.138     \\
        23$\times$23 & 1599760 & 215216 & 872162  & 2684.0 & 4.234     \\
        \hline
        \multicolumn{6}{|l|}{Maximum non-square grid}                  \\
        \hline
        24$\times$23 & 1698995 & 253580 & 918316  & 2684.0 & 5.639     \\ 
        \hline
        \multicolumn{6}{|l|}{Total available resources}                \\
        \hline
        \multicolumn{2}{|r}{1728000} & 791040 & 3456000 & 2688 & \\
        \hline
    \end{tabular}
    \vspace{-14pt}
\end{table}

In this section we quantitatively determine and discuss the FPGA scalability limits of the current RANC ecosystem.
All results are obtained through post implementation analysis using Vivado 2019.2.
Table~\ref{tab:alveo_scalability} presents the scalability for RANC's default configuration $N(a)=N(n)=256$ as a function of the resources available on the Alveo U250.
We find that, for this core configuration, BRAM availability is the limiting resource, and the design scales approximately linearly from 1 core up to the maximum supported grid of 552 cores.
With a 5.639ns critical path, this gives a core frequency of $f_{core} = 177.336$MHz, although many grids operate at much closer to 250MHz.
Given that 66,308 cycles are required for the \textit{\edit{core controller}} in the event that all 65,536 synapses are connected, this gives that the maximum global $f_{tick}$ is 2.674kHz.
However, by changing just the core sizing, we can cause LUTs to instead be the limiting factor.
To demonstrate this, we conducted another equivalent sweep where each RANC core was sized with $N(a)=N(n)=128$.
In this case, the BRAMs required per core drop from 5.5 to 3.5, and LUT resources are exhausted with scaling.
With this, we find that the maximum grid size is given by $26\times 26 = 676$ cores.
Small cores such as these may have benefit in applications that, i.e., require a high $f_{tick}$ rate as the internal clock rate for each core is $f_{core} = 241.31$MHz, and the number of cycles required by the \edit{core controller} in the worst case drops to 16,768.
Together, this gives an $f_{tick}$ of 14.39kHz.
If, instead, the goal is to emulate the largest possible architecture, we must determine the correct configuration that can maximally utilize all the FPGA fabric primitives.
As BRAM is the limiting resource to our scalability in the default configuration, that will be the focus here.
In our case, each baseline RANC core requires 5.5 72-bit$\times$512 word BRAM primitives.
As all BRAM primitives in the design are utilized by the \textit{\edit{core SRAM}}, we must ensure that we are utilizing all 512 words of the BRAM primitives and that each one of these words is $5.5\times 72 = 396$ bits in size.
The width of each word is easily controlled by the axon count, and the number of words is controlled by the number of neurons, so 
the core that maximizes utilization of a single set of BRAM primitives is a core with $N(a)=283$ and $N(n)=512$.
When we synthesize a single core with this configuration, we see that the resource utilization is unchanged outside of some logic expanding in width to deal with the larger \textit{\edit{core SRAM}} data words.
We then conducted a similar scalability analysis as previously, and we find, because of slight increase in LUT stress per core, that a slightly lower stopping point of 506 cores in a 23$\times$22 grid is entirely achievable.
In total, this configuration enables RANC to support 23$\times$22$\times$512$=$259,072 distinct neurons and 259,072$\times$283$=$73,317,376 distinct synapses.

%% file: sections_revised/5_related_work.tex
As the use cases for RANC intersect with a large number of research areas, we break the discussion on related work into two parts.
First, we discuss the large body of work in existing software simulators and discuss how they can be found to be complimentary to the goals of RANC.
Then, we discuss existing hardware architectures, and we note the differentiating factors that make RANC stand out.

Starting with software simulators, there has been a large amount of work in this area ranging from biologically accurate simulators for use in computational neuroscience to simulators inspired by the recent uptick in machine learning research.
On the biological front, NEST 2.20.0~\cite{fardet2020NEST},
Brian 2~\cite{stimberg2019brian2}, and NEURON~\cite{carnevale2006NEURON} are among the most widely known simulators. 
While the level of detail they provide is surely appreciated among neuroscientists, their bio-realism leads to long runtimes for those only interested in their application to computation.
Towards enabling enhanced usability across simulators, frameworks like PyNN~\cite{Davison2009PyNN} enable easy porting of models across all of these simulators.
CARLSim4~\cite{Chou2018CARLSim} is arguably suited for both biological and computational use cases, as while they initially started with a similar focus on GPU-accelerated, biologically plausible SNN simulation, they have begun to similarly apply CARLSim towards general neuromorphic computing applications through advances such as PyCARL~\cite{balaji2020pycarl}.
Finally, on the purely machine learning side, packages like SpykeTorch~\cite{mozafari2019spyketorch} are available to enable rapid training and testing of SNNs.
RANC differentiates itself from these simulation-oriented works by providing an end-to-end ecosystem that tightly couples software simulation and hardware emulation.
As such, we are not limited to simulation based experiments. 
Instead we can combine simulation and FPGA based analysis to progressively move from testing new ideas in software to implementing equivalent changes in hardware.
This allows rapid validation at the application level impacts through software and enables evaluating further impacts on latency, power, and resource utilization through hardware.
While frameworks like PyCARL~\cite{balaji2020pycarl} claim to have support for simulating commercial platforms such as TrueNorth and Loihi, to the best of our knowledge, no results are presented in this regard.
Additionally, the cycle-accurate simulator they use is a general purpose NoC simulator that requires precise SystemC implementations of the NoC tiles tested.
As far as we are aware, no SystemC implementations are available for either TrueNorth or Loihi.

Moving on to hardware platforms, most existing work in neuromorphic hardware has been the production of prefabricated ASICs in both the commercial sector with chips like Intel's Loihi~\cite{Davies2018Loihi}, IBM's TrueNorth~\cite{Akopyan2015TrueNorth}, the Akida Neuromorphic SoC~\cite{akida2019neuromorphic}\edit{, or Qualcomm Zeroth~\cite{qualcomm2013zeroth}} and the academic environment with chips like SpiNNaker~\cite{Painkras2012SpiNNaker}, DYNAPs~\cite{Moradi2018DYNAPs}, NeuronFlow~\cite{moreira_neuronflow_2020}\edit{, SpinalFlow~\cite{narayanan_spinalflow_2020}, or ODIN~\cite{frenkel_ODIN_2019}}.
\edit{
Of these, ODIN is notable in that it has an open source verilog implementation despite not being targeted towards FPGAs.
With that, it is similar to the RANC FPGA emulation environment, and it includes some features that RANC does not such as spike-driven synaptic plasticity (SDSP). 
Compared to RANC, however, ODIN has distinct goals with regards to energy minimization and biological plausibility. 
The ODIN chip includes only a single crossbar without a broader NoC that connects many parallel neuron crossbars together. 
As such, with ODIN, the focus is not on providing a scalable architecture for neuromorphic computing but instead maximizing energy savings and enabling large amounts of biologically realistic neuron behaviors. 
Without a broader NoC, the mapping methodologies required for applications like multilayer SNNs or VMM explored in this paper are quite unclear, and it cannot be used for the same style of architecture exploration as RANC’s focus on general purpose neuromorphic compute.
}
\edit{While RANC's FPGA emulation has a number of disadvantages with regards to power, latency, and resource usage compared to ASICs, t}he key advantage of RANC against \edit{ASIC} platforms such as these is that RANC is not a fixed architecture. 
To the contrary, RANC is built with architectural exploration in mind and as such it prioritizes easy exploration of novel methods of neuromorphic computing at the architectural level rather than only the application level.
If the production of improved ASIC architectures is the end goal, RANC positions itself as an environment where such architectural research can be performed.
In the vein of FPGA-based architectures, the closest academic platforms to what RANC offers are Minitaur~\cite{Neil2014Minitaur} or DANNA 2~\cite{Mitchell2018DANNA2}.

Minitaur is an FPGA-based SNN accelerator with an event-driven architecture.
Based on the details provided, the architecture presented takes a fundamentally different approach from the one taken by RANC.
Rather than for general neuromorphic computing, Minitaur is presented purely as an accelerator for spiking neural network inference, and while it is an FPGA-based design, the parameters given are presented as fixed to the design. 
As such, there are questions as to how flexible the design is in emulating network configurations with widely varied core requirements.
In particular, Minitaur appears to be less flexible in enabling general spike-based computation than the NoC approach utilized by RANC, as it is unclear how algorithms like vector matrix multiplication would be mapped.
Additionally, the fanin/fanout capabilities are wildly different, with Minitaur utilizing range-based destinations that map connections via start and end address ranges rather than directly mapping between neurons.
Finally, while the Minitaur design is quite compact, there are concerns with its throughput as, for instance, the MNIST results presented require 1000 spikes per image in order to achieve their maximum of 92\% accuracy on MNIST compared to the 1 to 16 spikes used in the studies presented here with 96+\% accuracy.
Whether this due to differences in network architecture, training methodology, or hardware is unclear.
Meanwhile, DANNA2 is the second iteration of DANNA~\cite{Disney2016DANNA}, an environment for neuromorphic computing that combines software simulation with FPGA and VLSI implementations.
Compared to RANC, the architecture presented in DANNA2 is quite different.
Each element in the DANNA2 grid has a fixed set of 24 input synapses that feed a single neuron unit, and this is overall much lower than a single RANC node that supports $N(a)$ inputs and $N(n)$ outputs per node. 
One consequence of this is that DANNA2 nodes inherently give a lower fanout per node than the $N(n)$ neurons in a single RANC core.
Coupled with the fact that each node only has a fixed set of 24 inputs compared to $N(a)$, many of the optimizations explored in VMM and SAR execution involving heterogeneous sizing of cores are infeasible on such an architecture.
For DANNA2 in particular, to the best of our knowledge, no published FPGA results are available to directly compare against.
This contrasts the published results for the original DANNA~\cite{Disney2016DANNA}, which depicts an even more divergent architecture.
Negative weights are unsupported and there are only 8 inputs per neuron, and consequently, concerns regarding the fanin/fanout capabilities of DANNA2 apply to DANNA as well.

\edit{
Finally, in the vein of hardware-software ecosystems for neuromorphic computing, two of the most relevant works are Nengo~\cite{Bekolay2014Nengo} and NAXT~\cite{abderrahmane_NAXT_2020}.
The Nengo ecosystem~\cite{Bekolay2014Nengo} provides a rich set of APIs to develop large-scale spiking and non-spiking neural networks on a wide range of platforms (e.g. FPGA, Loihi, HPC).
In particular, NengoFPGA requires a proprietary bitstream to interact with the Nengo APIs and does not provide a means to exploring novel neuromorphic architectures.
However, similar to NengoLoihi and NengoSpiNNaker, developing NengoRANC APIs in the future would enable co-design and execution of SNNs for user defined neuromorphic architectures in the RANC environment.
Meanwhile, NAXT~\cite{abderrahmane_NAXT_2020} provides an environment where users can generate FPGA-based SNN accelerators with a wide degree of flexibility in parameters of the generated architecture with regards to parallelism versus resource-usage tradeoffs. 
However, as NAXT is tailored specifically for SNN execution, the architectures it generates are specifically tailored for the baseline SNN architecture that is being mapped, and they are not suitable for use either by other SNNs with different architectures or non-machine learning applications in general. 
In this regard, RANC differentiates itself by providing architectures that are independent from the applications they execute and supporting non-machine learning applications that can still be decomposed to parallel spike-based computations like solving sparse coding via the locally competitive algorithm~\cite{fair19}.
}

\edit{
For the sake of completeness, we end by acknowledging some key limitations to the RANC framework as it currently stands and how the project’s goals align with these limitations. 
First, as our intention is neuromorphic computing over biologically realistic modeling, our ecosystem does not support features such as stochastic behavior or emulation of detailed neuronal dynamics, and we focus instead on the synaptic cores and a basic leaky-integrate-and-fire neuron model.
Second, while the architecture is amenable to this in the future by customizing the Router component, there is currently no support for NoC topologies other than the current 2D mesh. 
Finally, the core architecture is built around storing a dense-matrix form of synaptic connections, and supporting sparse alternatives will require architectural changes.
}
\edit{However, t}o the best of our knowledge, RANC is a wholly unique environment in its design goals and capabilities. 
While it has a software simulation component, it is distinct from existing software simulators through its ability to conduct wider studies impacting both hardware and software usage.
While other hardware designs have been proposed, RANC differs from existing ASICs by its reconfigurable nature and from FPGA implementations with its architectural differences and open source availability.

%% file: sections_revised/6_conclusion.tex
In this paper, we introduce RANC, an open source hardware and software ecosystem for neuromorphic computing.
We used the training, simulation, and emulation environments to validate functionality against existing baseline architectures and conduct rich hardware case studies on how improvements can be made beyond the baseline.
We illustrated how RANC allows the user to expose impacts of architectural design decisions through trend-based analysis and understand the impacts between different architectural parameters via combinatorial parameter sweeps.
With this level of functionality, we believe that these tools will stimulate further research into exploration of novel neuromorphic computing architectures.

As future work, we plan to incorporate richer architectural features such as multicast routing, on chip learning \edit{via spike timing dependent plasticity (STDP) and related learning methods}, and robust support for heterogeneous core configurations.
\edit{Learning via STDP variants in particular has been used to solve problems such as ECG classification~\cite{amirshahi_ecg_2019}, robot target tracking~\cite{bing_end_2019}, and network pruning~\cite{rathi_stdp-based_2019} quite successfully.}
\edit{
These added architectural features will extend RANC’s ability to emulate advanced neuromorphic architectures including, but not limited to, Intel’s Loihi.
As it stands, RANC has a number of existing features in common with Loihi. 
Like RANC, Loihi uses dimension-order routing to enable communication between the cores, each core receives and processes spike packets with LIF neurons, and these neurons themselves integrate received spikes to generate further spikes on the NoC. 
While the implementation details vary between the two platforms in how they manage core synchronization or particular features such as core multicast and neuron compartmentalization, the flexibility of RANC allows for gradual integration of differing aspects in the future.
}
Additionally, we will investigate how these features impact not only resource utilization but also the kinds of algorithms and applications that can be deployed on RANC.
Finally, to enable richer support for neural networks, we will incorporate support for other well-established Python-based SNN training packages to enable easily porting architectures from other environments.

%% file: sections_revised/9_appendices.tex
\renewcommand\thefigure{\thesection.\arabic{figure}}
\setcounter{figure}{0}
\renewcommand{\thetable}{\thesection.\Roman{table}}
\setcounter{table}{0}
\clearpage
\renewcommand{\thepage}{A-\arabic{page}}
\setcounter{page}{1}

\appendix[]

\subsection{Positive 4-bit VMM Execution} \label{app:vmm}

\begin{figure*}[!th]
    \centering
    \subfigure[]{
        \includegraphics[width=0.25\textwidth]{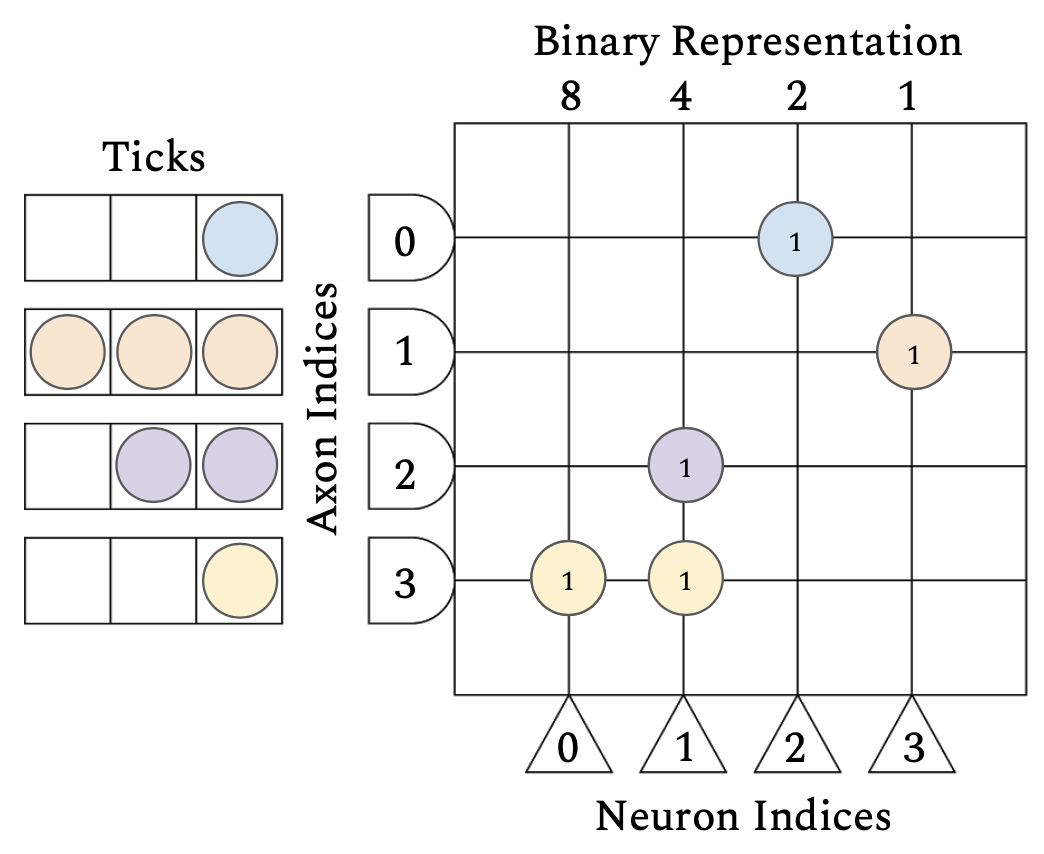}
        \label{fig:tik1}
    }
    \subfigure[]{
        \includegraphics[width=0.25\textwidth]{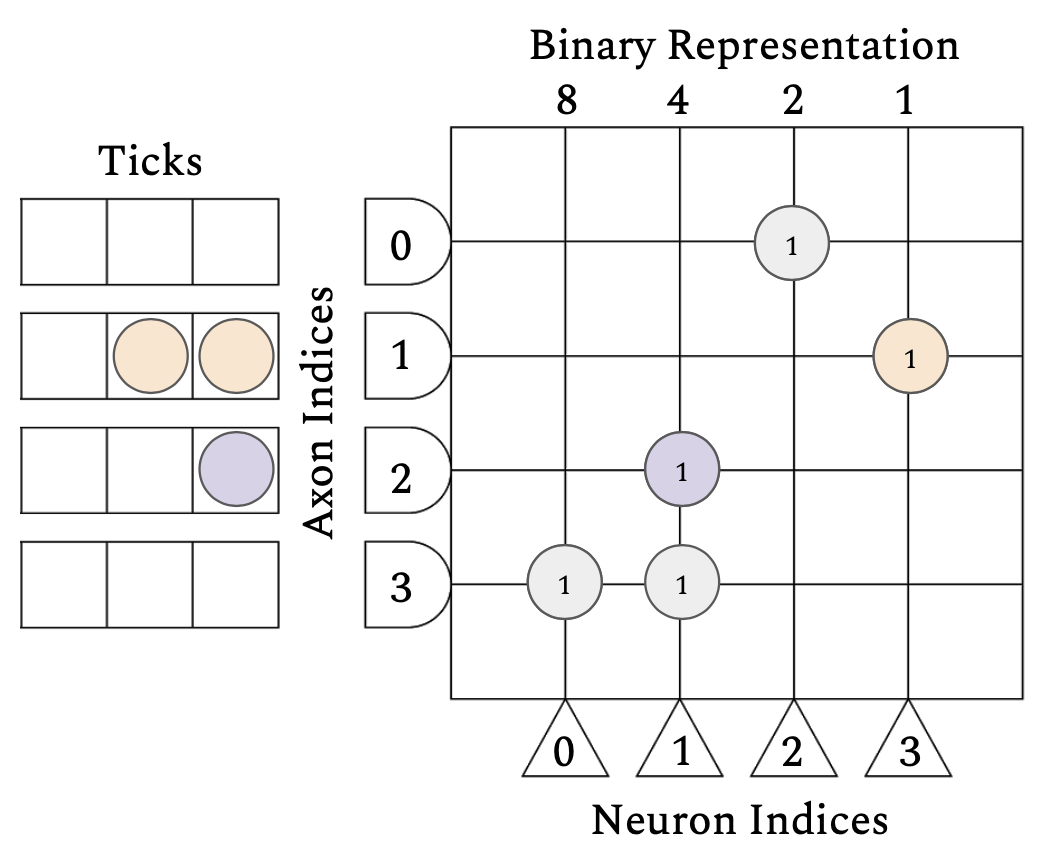}
        \label{fig:tik2}
    }
    \subfigure[]{
        \includegraphics[width=0.25\textwidth]{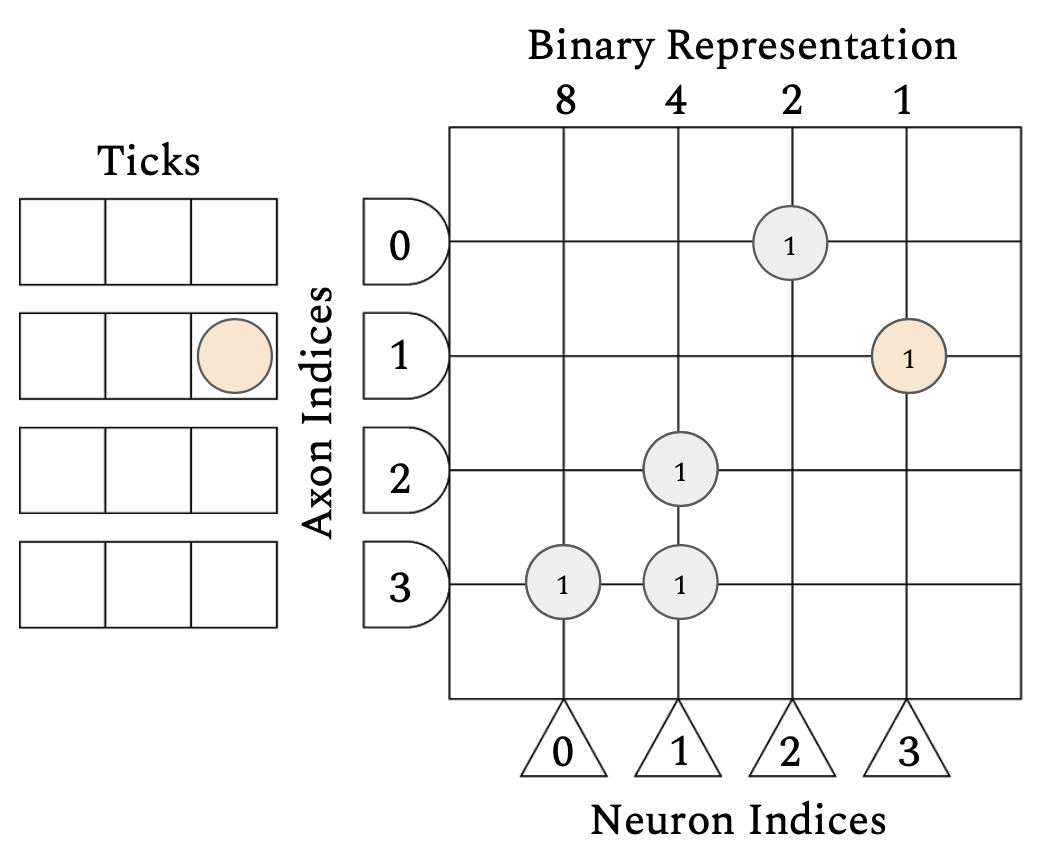}
        \label{fig:tik3}
    }
    \vspace{-2mm}
    \caption{Mapping of vector [1, 3, 2, 1] and matrix $[2, 1, 4, 12]^T$. Input vector is arranged vertically such that the number of arriving spikes equals the vector value at each axon. Matrix values are represented in a binary fashion where a connection and subsequent weight of 1 at a given neuron-axon intersection corresponds to the overlaid 4-bit mapping. Each tick permits a round of spikes into the crossbar, and sees the neuron potentials integrate accordingly. (a) shows the state at first tick, subsequent figures illustrate the state after each tick has elapsed.}
    \label{fig:core1mapping}
    \vspace{-16pt}
\end{figure*}

\begin{figure}[t]
    \centering
    \includegraphics[width=0.8\linewidth]{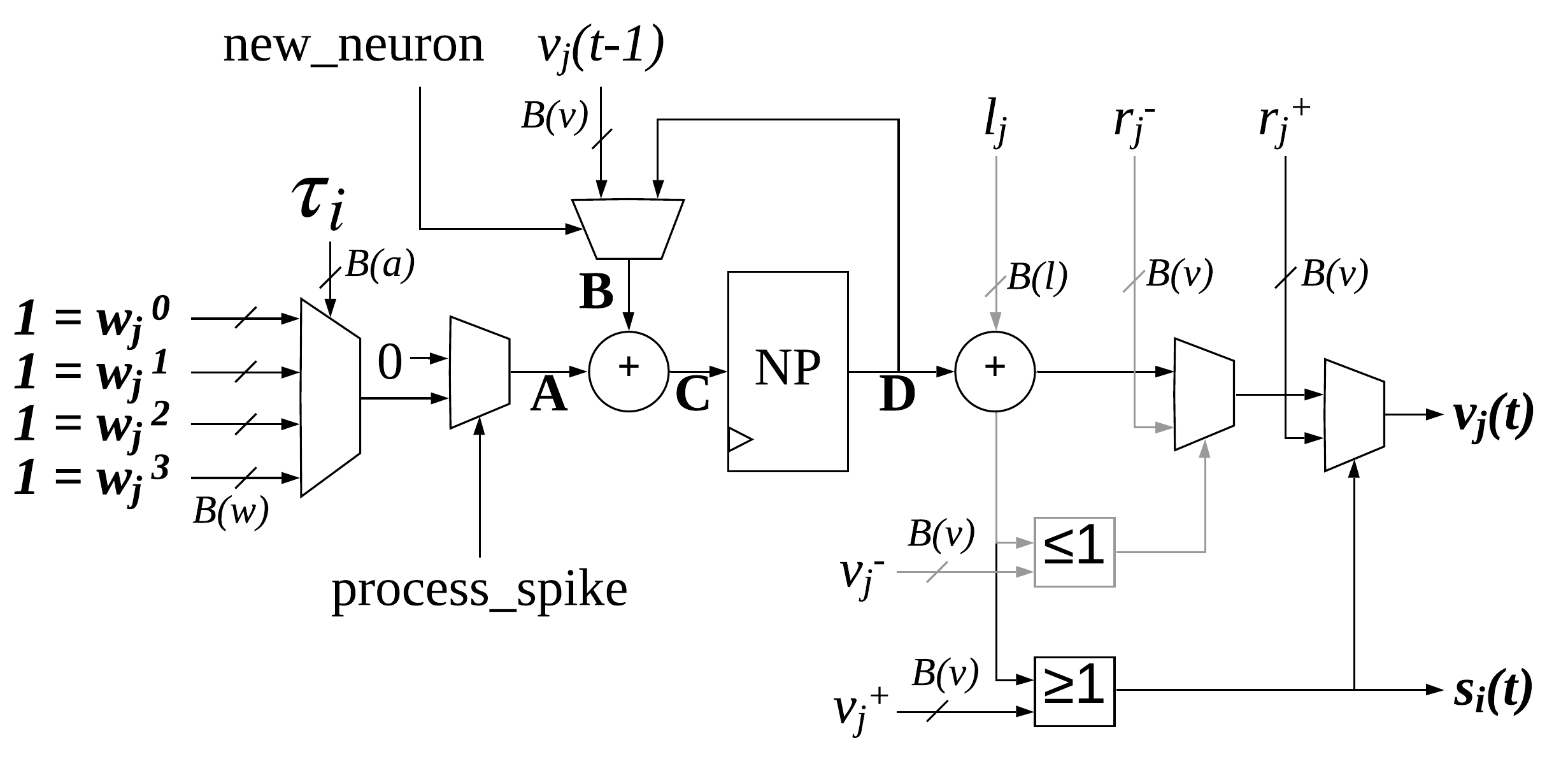}
    \vspace{-4mm}
    \caption{Neuron datapath's active components indicated with darker color for VMM implementation. Neuron weights are assigned as are required in core 1 of VMM.}
    \label{fig:nbvmm}
    \vspace{-5mm}
\end{figure}

\begin{figure}[t]
    \centering
    \includegraphics[width=0.4\linewidth]{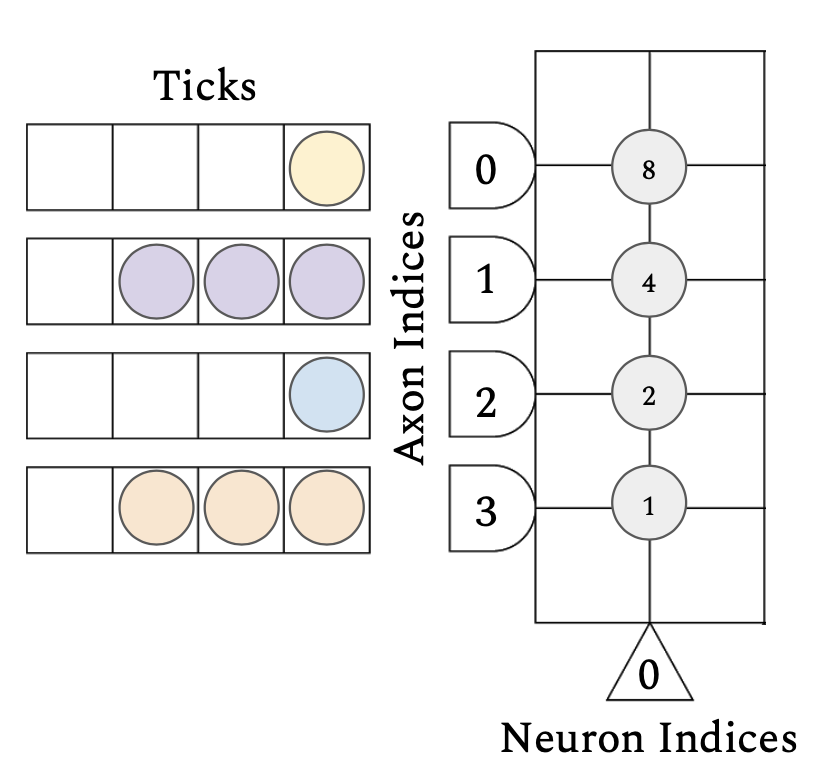}
    \vspace{-2mm}
    \caption{Core 2 of VMM applies integer weighting on Core 1's output. %
    }
    \label{fig:core2vmm}
\end{figure}

In this appendix, we illustrate the mapping methodology of positive VMM through example. 
The vector values in our example are $[1, 3, 2, 1]$ and the matrix values are $[2, 1, 4, 12]^T$. 
This particular VMM implementation requires two RANC cores. 
The first core performs the VMM as binary computation while the second core converts the binary representation to integer, producing a number of spikes correctly representing the final product.
In Figure~\ref{fig:core1mapping}, we illustrate the VMM implementation using a crossbar, which represents the computations in the first core.
In this crossbar, the \edit{$and$} shapes on the left represent axons ($a_i$) indexed from 0 to 3, arranged from top to bottom to align with the order of vector values. 
The \edit{$triangle$} shapes below the crossbar illustrate neurons ($n_j$), which are indexed from 0 to 3, arranged left to right.
The circles overlaid on the crossbar represent connection points and weights between axons and neurons. 
The four digits above the crossbar aid in illustrating that the connection points found in each row depict a binary representation of each matrix value. 
For example, the fourth row has connection points at the columns representing 8 and 4, which totals to 12, our fourth matrix value. 
On the left of Figure~\ref{fig:tik1}, each vector element ($[1,3,2,1]$) is converted to colored spike representation based on rate coding and is aligned to the respective axon from top to bottom. 
These spikes are delivered by the \emph{\edit{core controller}} from the \emph{packet scheduler} to the \emph{neuron \edit{datapath}}. 
Each square represents a tick of execution, and this particular example requires three ticks to process in the first core. 
We show state of the crossbar before starting each tick with Figures~A.\ref{fig:tik2}-\ref{fig:tik3}. 

To execute VMM, each neuron, shown in Figure~\ref{fig:nbvmm}, receives the axon type ($\tau_i$), synaptic weights (\edit{$w_{j}^{k}$}), previous neuron potential ($v_j(t-1)$), leak ($\ell_{j}$), positive threshold $v_{j}^{+}$ and axon connections from the \emph{\edit{core SRAM}} via the \emph{\edit{core controller}}. Each of the four participating neurons receives ($\tau_i$), equivalent to the axon index, from the \emph{\edit{core controller}}. This acts as the selector for the weight table mux, indexing the desired weight. 
Each axon type received from the \emph{\edit{core SRAM}} has a value of 1 indicating a binarized weight value for each existing connection. 
Figure~A.\ref{fig:tik1} illustrates the state of four neurons during the first tick and Table~\ref{tab:c0n1tiks} shows the tick by tick execution of the neuron $n_1$ with values observed on each labeled wire and register \edit{of the datapath} shown in Figure~\ref{fig:nbvmm}. 
\edit{In the following subsections, we describe the tick-by-tick behaviors observed by the first and second cores in this VMM execution.}

\begin{table}[t]
\centering
\tabcolsep=0.16cm
\caption{Tick by tick ($t$) state of Neuron 0 in Core 2 
within the [8,4,2,1] weighted core. Spikes ($S_i(t)$) output by $t=4$.
}
\label{tab:core2vmm}
\vspace{-2mm}
\begin{tabular}{|l|l|l|l|l|l|l|l|l|l|l|}
\hline
t & $a_j$ & $\tau_i$ & A & $v_j(t-1)$ & D  & NN & B  & C  & $v_j(t)$ & $s_i(t)$ \\ \hline
2 & 0  & 0  & 8 & \edit{0}       & 0  & 1  & 0  & 8  & \edit{0}     & \edit{0}     \\ \hline
2 & 1  & 1  & 4 & \edit{0}       & 8  & 0  & 8  & 12 & \edit{7}     & \edit{1}     \\ \hline
2 & 2  & 2  & 2 & \edit{0}       & 12 & 0  & 12 & 14 & \edit{11}     & \edit{1}     \\ \hline
2 & 3  & 3  & 1 & \edit{0}       & 14 & 0  & 14 & 15 & \edit{13}     & \edit{1}     \\ \hline
  & N(a)  &    &   &         &    &    &    &    & 14    & 1     \\ \hline
3 & 1  & 1  & 4 & 14      & 0  & 0  & 14 & 18 & \edit{0}     & \edit{0}     \\ \hline
3 & 3  & 3  & 1 & \edit{14}       & 18 & 0  & 18 & 19 & \edit{17}     & \edit{1}     \\ \hline
  & N(a)  &    &   &         &    &    &    &    & 18    & 1     \\ \hline
4 & 1  & 1  & 4 & 18      & 0  & 0  & 18 & 22 & \edit{0}     & \edit{0}     \\ \hline
4 & 3  & 3  & 1 & \edit{18}       & 22 & 0  & 22 & 23 & \edit{21}     & \edit{1}     \\ \hline
  & N(a)  &    &   &         &    &    &    &    & 22    & 1     \\ \hline
\end{tabular}
\vspace{-4mm}
\end{table}

\begin{table}[t]
\caption{Positive VMM mapping for Core 0, Neuron 1.}
\vspace{-2mm}
\label{tab:c0n1tiks}
\tabcolsep=0.18cm
\centering
\begin{tabular}{|l|l|l|l|l|l|l|l|l|l|l|}
\hline
t & $a_i$  & $\tau_i$ & A & $v_j(t-1)$ & D & NN & B & C & $v_j(t)$ & $s_i(t)$ \\ \hline
1 & 2      & 2        & 1 & \edit{0}          & 0 & 0  & 0 & 1 & \edit{0}        & \edit{0}     \\ \hline
1 & 3      & 3        & 1 & \edit{0}          & 1 & 0  & 1 & 2 & \edit{0}        & \edit{1}     \\ \hline
  & \edit{N(a)}    &          &   &            &   &    &   &   & 1        & 1     \\ \hline
2 & 2      & 0        & 1 & 1          & 0 & 0  & 1 & 2 & \edit{0}        & \edit{0}     \\ \hline
  & \edit{N(a)}    &          &   &            &   &    &   &   & 1        & 1     \\ \hline
\end{tabular}
\vspace{-4mm}
\end{table}

\subsubsection{Core 1}
We describe the tick by tick execution illustrated with  Table~\ref{tab:c0n1tiks} based on Figure~\ref{fig:nbvmm} and Figures~A.\ref{fig:tik2}-\ref{fig:tik3}. Figure~A.\ref{fig:tik1} illustrates the state of four neurons during the first tick and Table~\ref{tab:c0n1tiks} shows the tick by tick execution of the neuron $n_1$ with values observed on each labeled wire and register \edit{of the datapath} shown in Figure~\ref{fig:nbvmm}.

For the $n_1$ in Figure~A.\ref{fig:tik1}, no connection points exist for axons $a_0$ and $a_1$. Therefore axons $a_2$ and $a_3$ contribute to the potential and are evaluated sequentially starting with $a_2$. First the \edit{axon type} $\tau_i$ of 2 is retrieved from the \emph{\edit{core SRAM}} corresponding to $a_2$, and weight value of 1 (\edit{$w_{1}^{2}$}) is selected. This is integrated with an initial potential of zero received from the “neuron potential” register (NP), changing its value to 1. $n_1$ then processes $a_3$, receiving a ($\tau_3$) of 3 and selecting weight value (\edit{$w_{1}^{3}$}) of 1. This is integrated with the current NP value of 1, and updating its value to 2. Once all axons for a given neuron are processed, the current potential is evaluated against the "positive threshold", which is 1 in VMM mapping. Since "$\geq$1" condition is met, a spike $s_i(t)$ of 1 is generated by $n_1$ and the \emph{\edit{core controller}} delivers this to the router to be sent to the second core. When a spike of 1 is generated, a linear reset mode is needed~\cite{fair19} to ensure accurate final spike count. This is captured by the “positive reset value”, which is equivalent to the $NP-1$ and is output as $v_j(t)$ to be stored in \emph{\edit{core SRAM}} and retrieved as $v_j(t-1)$ in the next tick as the starting neuron potential. A similar process is executed for all neurons. Within the first tick of this example, $n_0$ integrates 1 spike, corresponding to the connection to $a_3$, $n_1$ integrates 2 spikes as it is connected to both $a_3$ and $a_4$, while $n_2$ and $n_3$ each integrate 1 spike, from $a_0$ and $a_1$ respectively.

During the second tick, illustrated with Figure~A.\ref{fig:tik2}, $a_2$ is the only contributor for $n_1$. Starting with the previous neuron potential of 1 retrieved from the \emph{\edit{core SRAM}} as $v_j(t-1)$ and integrating with the weight of 1 due to $a_2$, causing $n_1$ to generate another spike as illustrated in Table~\ref{tab:c0n1tiks}. During this tick, $n_3$ also participates and emits spike due to $a_1$. During the third tick, as seen in Figure~A.\ref{fig:tik3}, $n_3$ participates as only $a_1$ has a spike. This brings the total spikes emitted by each neuron to 1 for $n_0$, 3 for $n_1$, 1 for $n_2$, and 3 for $n_3$ over a total of 3 ticks.

\subsubsection{Core 2}
We describe the tick by tick execution illustrated with Table~\ref{tab:core2vmm} based on Figure~\ref{fig:core2vmm}. The axons of the second core receive the spikes from the neurons of the first core, where $n_j$ of first core attaches to $a_j$ of second core. The cores operate in a pipelined manner where during tick 1, the first core generates the first round of spikes and during tick 2, the second core starts processing these spikes. The spike packets of [1,3,1,3] generated by the first core are shown in Figure~\ref{fig:core2vmm} that illustrates the state of the second core. The tick by tick execution for $n_0$ in core two is illustrated in Table~\ref{tab:core2vmm}. At tick 2, all four axons starting with $a_0$ contribute and the values of 8, 4, 2, and 1 are accumulated in this order. At the end of this tick, the potential is evaluated at 15, far above the positive threshold of 1. This triggers the first spike to be emitted, and the application of the linear reset value of 1, decrementing the potential to 14. At tick 3, $v_j(t-1)=14$ is retrieved and integrated with the spikes on $a_1$ and $a_3$. These contribute 4 and 1 respectively, bringing the potential to 19. This is followed by a positive threshold decrement to 18 and subsequent second spike generation. At tick 4, $v_j(t-1)=18$ is added to the final round of input spikes from $a_1$ and $a_3$, producing 23. This is decremented by the positive threshold to 22, and the third spike is generated. For subsequent ticks, there are no incoming spikes to process on any axons. In this case, \texttt{process\_spike} is now zero, and the neuron datapath will integrate a zero with the previous potential. The neuron datapath continues to function by retrieving the previous potential, evaluating it at the positive threshold of 1, decrementing the potential by 1, and generating spikes as long as it is at or above the threshold. The remaining 22 spikes combined with the three spikes generated during the first three ticks result with a total of 25 spikes.

\subsection{Axon and Nueron Utilization for SAR Imagery}
\label{app:sar}
We present the details of hand analysis for the crossbar utilization for the \emph{Default} and \emph{Modified} RANC configurations. Within RANC's default configuration, the maximum kernel size is limited to 128 due to the axon count of 256, and therefore the 121 pixel kernel is the largest square (11x11) allowed per core. 
An 11x11 kernel striding over a 32x32 SAR image without padding results in a 22x22 map, or 484 individual strides. 
As we have two 11x11 kernels, and two neurons per core can be utilized, 484 cores are needed to conduct this convolution. 
For this default configuration, each pixel is replicated on average 60 times. Unique axon usage is at 1.7\%, and each core uses two neurons out of the 256 available, or 0.7\%. 
In our proposed modified architecture, we set $N(a)=1024, N(n)=256$. 1024 axons within a core allow for 512 pixels to be available for computation. In a similar fashion to the example within Figure \ref{fig:conv}, core axon size limits the maximum allowed kernel and image subsection available to a core to a 22x22 pixel region.

As the image size is 32x32, four separate 21x21 pixel regions supplied to four convolutional cores are sufficient to conduct the entire operation. 121 of the total required 484 stride operations are executed per core, resulting in 242 of 256 neurons used as each stride has a feature depth of two. The total output of this convolutional layer is 968 neurons, so neuron utilization increases to 94.5\%. The four cores use a total of 4096 axons, and since 2$\times$32$\times$32$=$2048 axons would be the ideal minimum required, effective axon utilization increases to 50\%.